\PassOptionsToPackage{table}{xcolor}
\documentclass{article} 
\usepackage{iclr2027_conference,times}

\usepackage{xspace}                          

\usepackage{xspace}

\newcommand{\ie}{\textit{i.e.},\xspace}

\newcommand{\query}{q}                                    
\newcommand{\numIdentities}{N}                            
\newcommand{\idSet}{\mathcal{I}}                          
\newcommand{\idIdx}{i}                                    
\newcommand{\auxIdIdx}{{\idIdx}'}                                    
\newcommand{\matchedIdx}{\hat{\idIdx}}
\newcommand{\exemplars}[1]{\mathcal{E}_{#1}}              
\newcommand{\exemplar}{e}                                 
\newcommand{\gallery}{\mathcal{G}}                        

\newcommand{\numModels}{K}                                
\newcommand{\modelIdx}{k}
\newcommand{\auxModelIdx}{k'}                              
\newcommand{\model}[1]{M_{#1}}                            
\newcommand{\modelSet}{\mathcal{M}}                       

\newcommand{\SimilarityMean}[1]{\bar s_{#1}}
\newcommand{\SimilarityStd}[1]{\rho_{#1}}
\newcommand{\StandardizedSimilarity}[3]{z_{#1}(#2,#3)}   
\newcommand{\FusedScoreMap}{F}
\newcommand{\FusedScore}[2]{\FusedScoreMap(#1,#2)}        
\newcommand{\matchScore}[3]{s_{#1}(#2, #3)}               
\newcommand{\fuseOp}[1]{\Phi_{#1}}                        
\newcommand{\idScore}[2]{S_{#1, #2}}                      

\newcommand{\CandidateCenter}[1]{m_{#1}}
\newcommand{\CandidateSpread}[1]{s_{#1}}
\newcommand{\PooledCenter}{\CandidateCenter{0}}
\newcommand{\PooledSpread}{\CandidateSpread{0}}
\newcommand{\CalibCount}[1]{n_{#1}}
\newcommand{\CalibCountRef}{\CalibCount{0}}
\newcommand{\CalibratedCenter}[1]{\hat m_{#1}}
\newcommand{\CalibratedSpread}[1]{\hat s_{#1}}
\newcommand{\CalibratedScoreMap}{\widetilde{\FusedScoreMap}}
\newcommand{\CalibratedScore}[2]{\CalibratedScoreMap(#1,#2)}

\newcommand{\CompositionWeight}{\lambda}
\newcommand{\CombinedScoreMap}{\FusedScoreMap_{\text{comp}}}
\newcommand{\CombinedScore}[2]{\CombinedScoreMap(#1,#2)}

\newcommand{\threshold}{\theta}

\newcommand{\GenuineTag}{\mathrm{gen}}
\newcommand{\ImpostorTag}{\mathrm{imp}}
\newcommand{\ScoreGroupMean}[3]{\mu^{#1}_{#2,#3}}         
\newcommand{\ScoreGroupVariance}[3]{(\sigma^{#1}_{#2,#3})^2}
\newcommand{\GenuineScoreMean}[2]{\ScoreGroupMean{\GenuineTag}{#1}{#2}}
\newcommand{\ImpostorScoreMean}[2]{\ScoreGroupMean{\ImpostorTag}{#1}{#2}}
\newcommand{\GenuineScoreVariance}[2]{\ScoreGroupVariance{\GenuineTag}{#1}{#2}}
\newcommand{\ImpostorScoreVariance}[2]{\ScoreGroupVariance{\ImpostorTag}{#1}{#2}}
\newcommand{\dpr}{d'}
\newcommand{\IdentitySeparation}[2]{\dpr_{#1,#2}}
\newcommand{\FusionWeights}[1]{\rvw^{#1}}
\newcommand{\FusionWeight}[2]{w^{#1}_{#2}}    
\newcommand{\QueryFusionWeight}[2]{w^{#1}_{#2}}          
\newcommand{\CombinedFusionWeight}[2]{\tilde{w}^{#1}_{#2}} 
\newcommand{\WeightSmoothing}{\alpha}
\newcommand{\WeightTemperature}{\tau}

\providecommand{\best}[1]{\textbf{#1}} \providecommand{\secondbest}[1]{\underline{#1}}
\newcommand{\ours}[1]{\textbf{#1}}
\newcommand{\ourslab}[1]{\ours{Ours (#1)}}
\newcommand{\pmm}{\ensuremath{\,{\scriptstyle\pm}\,}}
\definecolor{ourrowcolor}{HTML}{DCEBF7}   
\newcommand{\ourrow}{\rowcolor{ourrowcolor}}
\newcommand{\tightrules}{\setlength{\aboverulesep}{0pt}\setlength{\belowrulesep}{0pt}%
  \renewcommand{\arraystretch}{1.2}}

\definecolor{symface}{HTML}{F5A030}   
\definecolor{symcal}{HTML}{1FB0B0}    
\definecolor{symaim}{HTML}{2E86D6}    
\definecolor{symgait}{HTML}{3AA655}   
\definecolor{symreid}{HTML}{8E44AD}   
\definecolor{symins}{HTML}{C0392B}    
\newcommand{\eFace}{\textcolor{symface}{\ensuremath{\blacklozenge}}}
\newcommand{\eCAL}{\textcolor{symcal}{\ensuremath{\spadesuit}}}
\newcommand{\eAIM}{\textcolor{symaim}{\ensuremath{\blacksquare}}}
\newcommand{\eGait}{\textcolor{symgait}{\ensuremath{\blacktriangle}}}
\newcommand{\eReID}{\textcolor{symreid}{\ding{108}}}
\newcommand{\eReIDL}{\textcolor{symreid}{\ensuremath{\circledast}}}  
\newcommand{\eIns}{\textcolor{symins}{\ensuremath{\clubsuit}}}
\newcommand{\comb}[1]{\mbox{#1}}
\newlength{\combrowheight}
\newcommand{\combblock}[2]{%
  \multirow{-#1}{*}{\comb{#2}$\,\smash{\left\{
    \rule[\dimexpr-#1\combrowheight/2+2pt]{0pt}{\dimexpr#1\combrowheight-4pt}
  \right.}$}}

\usepackage{amsmath,amsfonts,bm}

\def\Figref#1{Figure~\ref{#1}}

\def\Tabref#1{Table~\ref{#1}}

\def\Secref#1{Section~\ref{#1}}

\def\eqref#1{(\ref{#1})}
\def\Eqref#1{Equation~\ref{#1}}

\def\1{\bm{1}}

\def\rvw{{\mathbf{w}}}

\DeclareMathAlphabet{\mathsfit}{\encodingdefault}{\sfdefault}{m}{sl}
\SetMathAlphabet{\mathsfit}{bold}{\encodingdefault}{\sfdefault}{bx}{n}

\usepackage{hyperref}
\usepackage{url}

\usepackage{amsmath,amssymb,amsthm}
\usepackage{graphicx}
\usepackage{booktabs}
\usepackage{multirow}
\usepackage[table]{xcolor}
\usepackage{subcaption}
\usepackage{pifont}

\theoremstyle{plain}

\theoremstyle{definition}

\title{Identity-Conditioned Score Fusion\\ for Open-Set Person Re-Identification}

\author{Manyi Yao$^{1}$\thanks{This work was done at Amazon.}\quad Jurijs Nazarovs$^{2}$\quad Eunji Chong$^{2}$\quad Abhishek Sharma$^{2}$\quad Rohan Sarkar$^{2}$\\
\bfseries Yue Guo$^{2}$\quad Christian R. Shelton$^{1}$\quad Amit Roy-Chowdhury$^{1}$\quad Debashish Pal$^{2}$\\
{\normalfont\small $^{1}$University of California, Riverside\qquad $^{2}$Amazon}\vspace{-0.15in}}

\iclrfinalcopy 
\begin{document}

\maketitle
\lhead{Preprint} 

\begin{abstract}
Robust person re-identification often combines complementary cues such as face, gait, and body shape. While adaptive fusion typically targets query quality, model strength also varies across identities. We introduce identity-conditioned score fusion, a framework that tailors weights to each gallery identity without training. By contrasting intra-identity consistency against cross-identity impostors, it extracts identity-specific profiles that couple with query-conditioned adaptation via a parameter-free rule. This widens the separation between true and false matches while preserving score calibration. Evaluations on three clothes-changing person re-identification benchmarks show that our method consistently outperforms statistical, rank-based, and learned baselines, achieving up to an 8.8\% absolute reduction in false non-identification rate and demonstrating the value of identity-conditioned fusion in open-set person re-identification.

\end{abstract}

\section{Introduction}
\label{sec:intro}

Person re-identification (ReID) benefits from models that capture complementary biometric cues, including face, gait, and body shape~\citep{farsight2024,liu2025person}. Their usefulness varies with the observations: face cues become less reliable at low resolution or oblique angles, while gait requires a suitable video sequence. Combining these models therefore calls for a fusion rule that determines how much each should contribute to a match.

Recent fusion methods address variation in cue availability by adapting to the current inputs~\citep{gupta2020quality,qme2025,hamobe2025,fusionagent2026}. Quality estimates, learned gates, and model-selection policies adjust which models contribute and how their evidence is combined. Such adaptation can reduce reliance on a face model when the face is poorly visible or on another model whose required cues are degraded. It accounts for changes in the observations, but a usable cue need not be equally discriminative for every enrolled person.

A model's contribution also depends on how well it distinguishes the identity being matched from other people in the gallery. Similar embeddings across that identity's samples are useful only if the model also separates them from samples of other identities. A model that clearly separates one person from the rest of the gallery may provide little separation for another. Assigning the same relative model weights to every gallery identity cannot express this difference. Classical user-dependent biometric fusion has investigated such variation across individuals~\citep{snelick2005user,fierrez2005bayesian,poh2013user}. The labeled gallery provides evidence for estimating these differences through exemplars of each person and comparisons with other enrolled identities.
\begin{figure}[t]
\centering
\includegraphics[width=.7\linewidth]{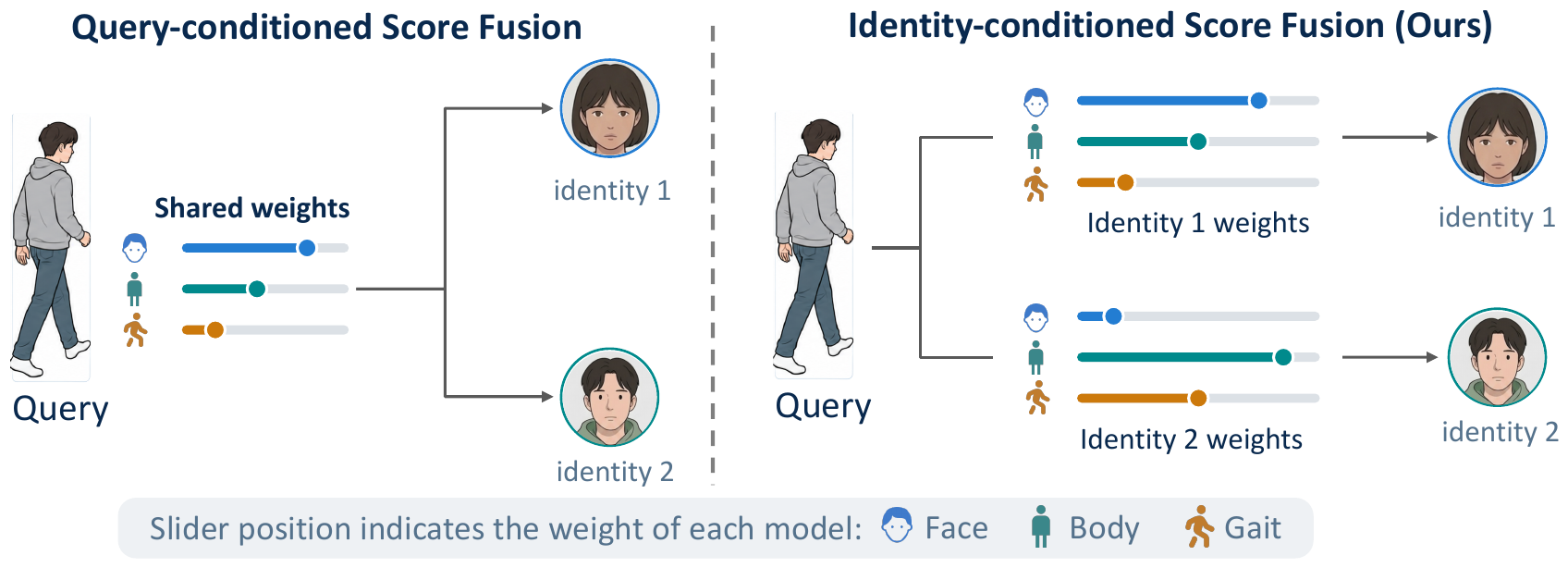}
\caption{\textbf{Query-conditioned vs. Identity-conditioned score fusion.} Existing multi-modal gait recognition often applies a global, shared weight to all identities (left). In contrast, our proposed Identity-conditioned approach adaptively tailors weights for each identity profile (right), allowing the system to leverage the most reliable modality on an identity-specific basis. Sliders conceptually illustrate the weight assignment.}
\label{fig:teaser}
\end{figure}

We introduce identity-conditioned score fusion, a method that uses these gallery comparisons to estimate a model-weight profile for each enrolled identity. For each model, similarities within an identity measure consistency across its reference samples, while similarities to all other gallery identities indicate how readily other people could be confused with it. The separation between these two groups determines the model's weight. At query time, matching scores computed from the models' embeddings are combined using the profile of the gallery identity being considered. The same query can therefore be compared with different identities using different model weights, as illustrated in \Figref{fig:teaser}. The profiles are reused across queries without updating the recognition models or training an additional fusion network. For open-set decisions, calibration accounts for differences in non-match scores across identities. Composition with query-dependent fusion allows the weights to be used alongside adaptation to the current observation.

Experiments on three person re-identification benchmarks demonstrate gains over competing fusion methods using the same underlying models, both with identity-conditioned fusion alone and in composition with query-dependent methods. Further analysis shows that different identities benefit from different models, and that gallery comparisons help identify these preferences. We also study how these gains depend on the exemplar counts used to estimate the weights.
Our contributions are:
\begin{itemize}
    \item We propose an identity-conditioned score fusion method for open-set person ReID that dynamically tailors model weights to enrolled gallery profiles without training.
    \item We show that our method seamlessly unifies with query-conditioned fusion via a parameter-free combination rule, enabling them to jointly leverage complementary cues.
    \item Extensive evaluations on three benchmarks demonstrate that our method consistently outperforms statistical, rank-based, and learned fusion baselines, effectively widening the genuine--impostor separation margin.
\end{itemize}

\section{Related Work}
\label{sec:related}

\paragraph{Open-set person re-identification.}
Open-set person ReID requires matching queries to enrolled identities and rejecting queries whose identities are absent from the gallery~\citep{liao2014openset,zheng2016openworld,qme2025}.
Robust matching across appearance changes has motivated representations that reduce clothing dependence~\citep{prcc2021,cal2022,aim2023,differ2025,icsm2025,csci2025,sapiensid2025} and exploit body shape or gait~\citep{invarreid2023,clip3dreid2024,ccpg2023,biggait2024,seas2024,gaitmax2026}.
Reliable rejection also requires separating true matches from high-scoring non-matches~\citep{openset2024,localscore2026}.
Open-set biometric studies show that good closed-set performance does not necessarily yield reliable rejection and introduce training losses that explicitly account for impostor scores as well as correct identity matches~\citep{openset2024,qme2025}.

\paragraph{Adaptive multimodal fusion.}
Combining face, body shape, and gait can compensate for unreliable observations in individual modalities~\citep{farsight2024,liu2025person}.
Score-level fusion combines outputs from independently trained models~\citep{ross2003fusion,nandakumar2008likelihood,qme2025}, with normalization accounting for differences in score scale~\citep{snelick2003multimodal,jain2005score,he2010performance,fusionagent2026}.
Quality-dependent fusion further adjusts the contribution of each modality according to the reliability of the observed input~\citep{nandakumar2006quality,poh2009benchmarking,poh2012unified,gupta2020quality,pdf2024,cocoon2025,evidencefilter2025}.
QME~\citep{qme2025} combines learned fusion strategies according to estimated query quality, while FusionAgent~\citep{fusionagent2026} selects models for each query before confidence-based score fusion.
HAMoBE~\citep{hamobe2025} conditions feature fusion on both videos in a query--gallery pair, allowing the feature weights for a query to vary with the gallery video being compared.

\paragraph{Identity-specific biometric fusion.}
Biometric matchers exhibit identity-dependent performance~\citep{doddington1998sheep,yager2010menagerie}, which inspired early verification studies to weight classifiers using client-specific separation statistics~\citep{snelick2005user,poh2013user}. However, these proportional weighting rules rely on an assumption that separation measured at enrollment remains valid at test time. In person Re-ID, camera and viewpoint changes frequently break this assumption, making rigid proportional weights overconfident. We instead formulate a regularized weighting strategy that mitigates these transfer errors while integrating model consensus directly into the identity scoring pipeline.
\section{Proposed Method}
\label{sec:method}

\begin{figure}[t]
\centering
\includegraphics[width=.97\linewidth]{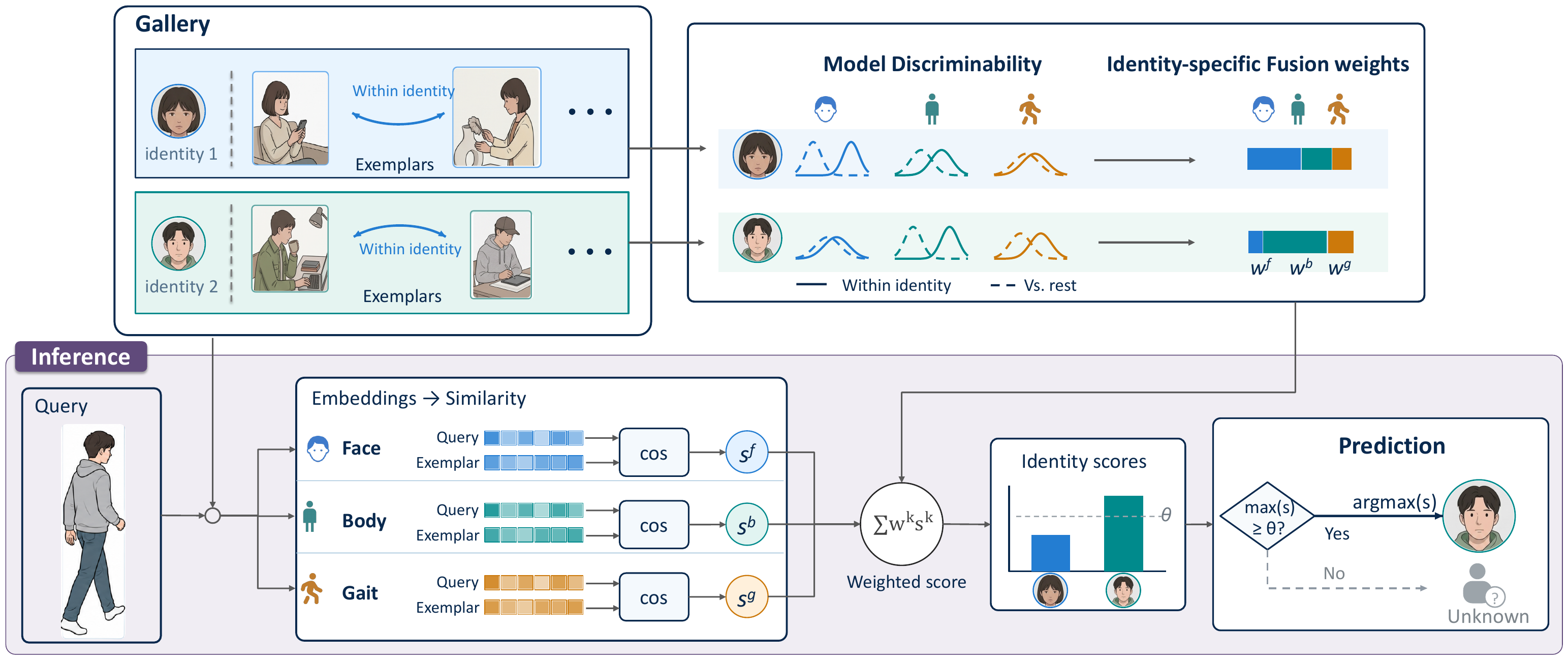}
\caption{\textbf{Overview of our framework.} \textit{Top:} Our method leverages gallery intra- and inter-identity statistics to derive tailored model reliability weights $\FusionWeight{\idIdx}{\modelIdx}$ for each identity $\idIdx$ without auxiliary training.
\textit{Bottom:} When an incoming query is compared against the gallery, its similarity scores with candidate identity $\idIdx$ across modalities are combined using that candidate's specific profile, prioritizing reliable modalities for each individual for final open-set verification.}
\label{fig:overview}
\end{figure}

\paragraph{Problem Setup}
\label{sec:method:setup}

Open-set biometric identification matches a query $\query$ (an image or video) against a gallery $\gallery$ of $\numIdentities$ enrolled identities, assigning it to one of them or rejecting it as unenrolled. Each identity $\idIdx \in \idSet = \{1, \ldots, \numIdentities\}$ is represented by a set of gallery exemplars $\exemplars{\idIdx}$, forming the gallery $\gallery = \bigcup_{i \in \idSet} \exemplars{\idIdx}$. 
Consider $\numModels$ fixed, pretrained models $\modelSet = \{\model{1}, \ldots, \model{\numModels}\}$, each mapping visual inputs to embeddings that capture biometric cues such as face or gait. A matching function, such as cosine similarity, compares the embeddings produced by model $\model{\modelIdx} \in \modelSet$ for query $\query$ and gallery exemplar $\exemplar \in \gallery$, yielding a matching score $\matchScore{\modelIdx}{\model{\modelIdx}(\query)}{\model{\modelIdx}(\exemplar)}$.
For each candidate identity $\idIdx$, a fusion operator $\fuseOp{\idIdx}$ aggregates these cross-model, multi-exemplar scores into an overall confidence score $\idScore{\query}{\idIdx} \in \mathbb{R}$ as
\begin{equation}
\label{eq:score_fusion}
\idScore{\query}{\idIdx} = \fuseOp{\idIdx}\Big( \big\{ \matchScore{\modelIdx}{\model{\modelIdx}(\query)}{\model{\modelIdx}(\exemplar)} \;\big|\; \model{\modelIdx} \in \modelSet, \; \exemplar \in \exemplars{\idIdx} \big\} \Big).
\end{equation}
Open-set identification assigns an identity only when an acceptance criterion is met; otherwise, the query is rejected as unenrolled. A common rule accepts the highest-scoring identity $\arg\max_{\idIdx \in \idSet}\idScore{\query}{\idIdx}$ if its score exceed a threshold. Closed-set one selects the highest-scoring identity without rejection.


As illustrated in \Figref{fig:overview}, we first estimate a mixture profile for each gallery identity from its gallery comparisons in \Secref{sec:method:profiles}). These profiles determine the score-fusion rule and its open-set decisions \Secref{sec:method:fusion}. We then describe their composition with query-dependent fusion in \Secref{sec:method:composition}.

\subsection{Gallery-derived Mixture Profiles}
\label{sec:method:profiles}

A biometric model may distinguish some enrolled identities more reliably than others. To determine a model's contribution for a candidate identity $\idIdx$, we quantify how effectively it separates identity $\idIdx$ from the rest of the gallery. Specifically, we evaluate two factors using the reference observations in the gallery: intra-identity similarity, which reflects consistency across observations of the same identity, and inter-identity similarity, which measures distinctiveness against others. An informative model should exhibit high intra-identity similarity and low inter-identity similarity, whereas a model assigning high scores to both groups offers little discriminative power. Intuitively, this separation margin reflects a model's reliability and governs its contribution.

\paragraph{Estimating identity-specific separation.}
To estimate the separation for identity $\idIdx$ under model $k$, we compute two sets of cosine similarities: the \emph{genuine} scores between distinct samples in its exemplars set $\exemplars{\idIdx}$ (capturing consistency), and the \emph{impostor} scores comparing samples in $\exemplars{\idIdx}$ with $\gallery\setminus\exemplars{\idIdx}$ (capturing distinctiveness). Both sets are conditioned on the target identity. The difference between the genuine and impostor means measures their average separation, but the variation within each group determines how informative that difference is. 
Following separation-based user weighting \citep{snelick2005user}, we quantify this margin using the signal-detection sensitivity index $\dpr$ \citep{green1966signal}, which scales the difference in means by their pooled standard deviation. Letting $(\GenuineScoreMean{\idIdx}{\modelIdx}, \GenuineScoreVariance{\idIdx}{\modelIdx})$ and $(\ImpostorScoreMean{\idIdx}{\modelIdx}, \ImpostorScoreVariance{\idIdx}{\modelIdx})$ denote the empirical mean and variance of the genuine and impostor scores, respectively, the separation achieved by model $k$ is defined as
\begin{equation}
\label{eq:dprime}
\IdentitySeparation{\idIdx}{\modelIdx}=\frac{\GenuineScoreMean{\idIdx}{\modelIdx}-\ImpostorScoreMean{\idIdx}{\modelIdx}}
{\sqrt{\tfrac12\bigl(\GenuineScoreVariance{\idIdx}{\modelIdx}+\ImpostorScoreVariance{\idIdx}{\modelIdx}\bigr)}}.
\end{equation}
Higher values of $\IdentitySeparation{\idIdx}{\modelIdx}$ reflect greater discriminability for identity $\idIdx$ within the gallery. By evaluating models independently, this metric avoids the instability of estimating cross-model covariance matrices under scarce gallery samples, while preserving modularity when scaling to larger ensembles.

Given the resulting separation scores across all $\numModels$ models, $\{\IdentitySeparation{\idIdx}{\modelIdx}\}_{\modelIdx=1}^{\numModels}$, a straightforward approach would be to map them into normalized weights via softmax, favoring models with higher discriminability. However, when gallery observations are limited, empirical separation estimates carry noticeable variance, and a standard softmax can yield an overly peaked distribution that heavily over-commits to a single dominant model. To hedge against this uncertainty, we regularize the allocation by interpolating the softmax distribution with a uniform prior across all $\numModels$ models, forming identity $\idIdx$'s mixture profile $\FusionWeights{\idIdx} = [\FusionWeight{\idIdx}{1}, \ldots, \FusionWeight{\idIdx}{\numModels}]^\top$ with each component defined as
\begin{equation}
\label{eq:softmax}
\FusionWeight{\idIdx}{\modelIdx} = \frac{1-\WeightSmoothing}{\numModels} + \WeightSmoothing \frac{\exp(\IdentitySeparation{\idIdx}{\modelIdx}/\WeightTemperature)}{\sum_{\auxModelIdx=1}^{\numModels}\exp(\IdentitySeparation{\idIdx}{\auxModelIdx}/\WeightTemperature)}\;,
\end{equation}
where $\WeightTemperature > 0$ controls the sharpness of the separation preference, and $\WeightSmoothing \in [0,1]$ balances data-driven weighting against uniform model averaging. This formulation prevents any single model from being prematurely discounted. Importantly, this mixture profile is computed directly from static gallery comparisons in closed form, requiring no additional training of a fusion network.

\subsection{Identity-conditioned Score Fusion}
\label{sec:method:fusion}
With the identity-specific mixture profile $\FusionWeights{\idIdx}$ established, we fuse multi-model predictions for an incoming query $\query$ against candidate identity $\idIdx$. Directly taking a weighted sum of raw cosine scores, however, is prone to distortion because individual models inherently operate across disparate score distributions and dynamic ranges; a model with numerically dominant scores would override the calibrated mixture weights. To place all models on a shared scale, we map each raw similarity $\matchScore{\modelIdx}{\model{\modelIdx}(\query)}{\model{\modelIdx}(\exemplar)}$ into a standardized score as
\begin{equation}
\label{eq:score_normalization}
\StandardizedSimilarity{\modelIdx}{\query}{\exemplar}=\frac{\matchScore{\modelIdx}{\model{\modelIdx}(\query)}{\model{\modelIdx}(\exemplar)}-\SimilarityMean{\modelIdx}}{\SimilarityStd{\modelIdx}},
\end{equation}

where $\SimilarityMean{\modelIdx}$ and $\SimilarityStd{\modelIdx}$ represent the empirical score mean and stabilized standard deviation for model $\modelIdx$ across the evaluation set. These normalization parameters are identity-agnostic and computed once per model without query labels.

Next, we combine these standardized scores to evaluate each candidate identity. Following standard practice in exemplar-based person Re-ID, we compute an ensemble consensus for each reference sample before aggregating across the identity's exemplar set $\exemplars{\idIdx}$, defining the uncalibrated fused score $\FusedScore{\query}{\idIdx}$ between query $\query$ and candidate identity $\idIdx$ as
\begin{equation}
\label{eq:fusion}
\FusedScore{\query}{\idIdx}=\max_{\exemplar\in\exemplars{\idIdx}}\sum_{\modelIdx=1}^{\numModels}\FusionWeight{\idIdx}{\modelIdx},\StandardizedSimilarity{\modelIdx}{\query}{\exemplar}\;.
\end{equation}
Under this formulation, multi-model predictions are fused per exemplar prior to max-pooling over $\exemplars{\idIdx}$, ensuring that the aggregate score reflects the strongest reference match without being diluted by less informative observations.

While this fused score $\FusedScore{\query}{\idIdx}$ provides an effective ranking criterion for closed-set retrieval, open-set person Re-ID requires deciding whether to accept the top candidate or reject the query as an unseen identity. This decision relies on a shared threshold; however, uncalibrated scores are susceptible to identity-specific baseline shifts, where certain candidate identities systematically elicit higher similarity from non-matching inputs. Using a uniform threshold directly on $\FusedScore{\query}{\idIdx}$ would disproportionately produce false acceptances for these intrinsically high-scoring identities.

\paragraph{Open-set calibration.} 
\label{sec:method:calibration}
To align candidate scores onto a commensurable scale, we adapt score normalization principles from biometric verification~\citep{auckenthaler2000score} to calibrate each identity against its own non-match distribution. Treating each non-matching gallery sample as a pseudo-query yields an empirical impostor score set $\{\FusedScore{\exemplar}{\idIdx}\;\big|\;\exemplar \in \gallery \setminus \exemplars{\idIdx}\}$. 
As established in extreme value analysis for open-set recognition~\citep{scheirer2013toward,bendale2016towards}, false acceptances are governed by extreme non-match outliers rather than central impostor tendencies. We therefore isolate the most competitive impostors by estimating candidate-specific tail statistics $(\CandidateCenter{\idIdx}, \CandidateSpread{\idIdx})$, defined as the empirical mean and standard deviation of impostor scores exceeding the 90th percentile. Because tail parameters estimated from finite gallery observations can exhibit high sample variance, we regularize them via empirical Bayes shrinkage, pulling the local estimates toward the pooled impostor statistics $(\PooledCenter, \PooledSpread)$ across the full gallery to obtain the calibrated mean $\CalibratedCenter{\idIdx}$ and standard deviation $\CalibratedSpread{\idIdx}$ as
\begin{equation}
\label{eq:calibration_estimates}
\CalibratedCenter{\idIdx}=\frac{\CalibCount{\idIdx} \CandidateCenter{\idIdx}+\CalibCountRef \PooledCenter}{\CalibCount{\idIdx}+\CalibCountRef}\qquad\text{and}\qquad\CalibratedSpread{\idIdx}=\frac{\CalibCount{\idIdx} \CandidateSpread{\idIdx}+\CalibCountRef \PooledSpread}{\CalibCount{\idIdx}+\CalibCountRef}\;,
\end{equation}
where $\CalibCount{\idIdx}$ denotes the number of valid impostor comparisons for identity $\idIdx$, and the reference count $\CalibCountRef$ is the median of positive comparison counts across all enrolled identities. Using these regularized statistics, the calibrated score is computed via standard z-normalization as
\begin{equation}
\label{eq:calib}
\CalibratedScore{\query}{\idIdx}=\frac{\FusedScore{\query}{\idIdx}-\CalibratedCenter{\idIdx}}{\CalibratedSpread{\idIdx}}\;.
\end{equation}
During open-set person re-identification, the calibrated score serves as the final identification score, \ie $\idScore{\query}{\idIdx} = \CalibratedScore{\query}{\idIdx}$. A query $\query$ is assigned to the top candidate identity $\matchedIdx = \arg\max_{\idIdx} \idScore{\query}{\idIdx}$ whenever its score $\idScore{\query}{\matchedIdx}$ exceeds a shared decision threshold $\threshold$, and is rejected as an unseen identity otherwise.

\subsection{Composition with Query-conditioned Fusion}
\label{sec:method:composition}
The identity-conditioned weights $\FusionWeights{\idIdx}$ are derived entirely from the given gallery $\gallery$, characterizing which models best isolate candidate $\idIdx$ from the remaining enrolled identities. While this profile captures intrinsic identity discriminability, it remains agnostic to the specific observation conditions of an incoming query $\query$ at inference time. Even if a model generally provides high discriminability for an identity, its reliability drops if that specific modality is degraded in the incoming query.
Query-conditioned fusion addresses this blind spot by assessing how trustworthy each model is on the observed input. Thus, identity-conditioned fusion and query-dependent fusion rely on fundamentally different signals: gallery-side separability identifies which features make a person distinctive, while query-side gating assesses which features are clean enough to trust. Fusing these predictions lets the final decision rule account for both candidate uniqueness and query condition.

Let $\FusionWeight{\query}{\modelIdx}$ denote the dynamic weight assigned to modality $\modelIdx$ by an arbitrary query-conditioned fusion rule, such as QME \citep{qme2025} or FusionAgent \citep{fusionagent2026}. Rather than interpolating scores, we integrate query-level visual quality with our identity-level distinctiveness directly at the weight level via element-wise multiplication. For query $\query$ and candidate identity $\idIdx$, the composite weight for modality $\modelIdx$ is defined as
\begin{equation}
\label{eq:compose_weight}
\CombinedFusionWeight{\query}{\idIdx}{\modelIdx}
= \frac{\FusionWeight{\idIdx}{\modelIdx} \cdot \QueryFusionWeight{\query}{\modelIdx}}{\sum_{\modelIdx'} \FusionWeight{\idIdx}{\modelIdx'} \cdot \QueryFusionWeight{\query}{\modelIdx'}}\;,
\end{equation}
which acts as a joint gating mechanism: a modality contributes strongly only when it is both inherently discriminative for identity $\idIdx$ and sufficiently reliable in query $\query$. The composite candidate score is then computed as $\CombinedScore{\query}{\idIdx} = \sum_{\modelIdx} \CombinedFusionWeight{\idIdx}{\modelIdx} \matchScore{\modelIdx}{\query}{\idIdx}$, which directly replaces $\FusedScore{\query}{\idIdx}$ in the calibration pipeline in \Eqref{eq:calib}. We evaluate identification performance both with and without this composition in \Secref{sec:exp}.
\section{Experiments}
\label{sec:exp}
\paragraph{Datasets and Evaluation Protocol.}
To rigorously validate the effectiveness and robustness of our proposed method, we evaluate it on three widely used clothes-changing person re-identification benchmarks: CCVID \citep{ccvid2022} and MEVID \citep{mevid2023} contain video sequences, while LTCC \citep{ltcc2020} contains images; detailed statistics are summarized in \Tabref{tab:datasets}. Following standard evaluation practices \citep{qme2025}, performance is measured using Rank-1 accuracy and mean Average Precision (mAP) for closed-set ranking, alongside two standard whole-body biometric metrics: TAR@1\%FAR (true accept rate at a $1\%$ false accept rate) for verification, and FNIR@1\%FPIR ($\downarrow$) for open-set identification \citep{grother2019frvt}. Following \citep{qme2025, fusionagent2026}, while Rank-1, mAP, and TAR are computed over the full gallery, FNIR is averaged across 10 runs where random  $80\%$ of gallery identities serve as mates and the remaining $20\%$ as non-mates to reflect unconstrained open-set scenarios.

\begin{table}[t]
\centering
\caption{\textbf{Statistics of person re-identification benchmarks.} 
Query and gallery counts refer to video sequences for CCVID \citep{ccvid2022} and MEVID \citep{mevid2023}, and images for LTCC \citep{ltcc2020}.
Identity counts are reported as train/test. The mean and range of gallery exemplar counts are computed across gallery identities in the test split.}
\label{tab:datasets}
\small
\setlength{\tabcolsep}{5pt}
\tightrules
\begin{tabular}{llccccc}
\toprule
\multirow{2}{*}{Dataset} & \multirow{2}{*}{Type}
  & \# Identities & \multirow{2}{*}{\# Queries}
  & \multicolumn{3}{c}{\# Gallery exemplars} \\
\cmidrule(lr){5-7}
  & & Train / Test & & Total & Mean / identity & Range / identity \\
\midrule
CCVID & Video & 75 / 151 & 834 & 1,074 & 7.1  & 3--12 \\
MEVID & Video & 104 / 54 & 316 & 1,438 & 26.6 & 1--101 \\
LTCC  & Image & 77 / 75  & 493 & 7,050 & 94.0 & 4--589 \\
\bottomrule
\end{tabular}
\end{table}

\paragraph{Recognition models.}
To leverage the complementary strengths of global appearance and localized biometric traits, we evaluate heterogeneous combinations of face, body, and gait recognition backbones (AdaFace~\citep{adaface2022}, CAL~\citep{cal2022}, BigGait~\citep{biggait2024}, AGRL~\citep{agrl}, and InsightFace/ArcFace~\citep{arcface2019}; specified in Tables~\ref{tab:ccvid_mevid} and~\ref{tab:ltcc}). For body representation, CLIP-ReID backbones~\citep{clipreid2023} are evaluated either zero-shot to test generalization or with LoRA adaptation~\citep{lora2022} on target training splits. Each model yields an independent similarity score channel. All upstream recognition models remain fixed during fusion to isolate the contribution of our fusion strategy from backbone fine-tuning. Full model details are provided in Appendix~\Secref{app:model_pools}.

\paragraph{Baseline setup.}
To fairly assess performance across different levels of supervision, we benchmark against two broad paradigms using matched recognition backbones:
(1) \emph{training-free baselines} that require no fusion training set, including fixed rules (Min/Max~\citep{jain2005score}, Z-score/Min-Max normalization~\citep{snelick2003multimodal}, FarSight~\citep{liu2025person}) and score-distribution alignment (RHE~\citep{he2010performance}); and
(2) \emph{training-reliant methods} optimized on the training split, including parameter-fitted rules (Asym-AO~\citep{herbadji2020combining}, grid-searched Weighted-sum~\citep{park2021weighted}), learning-based classifiers (BSSF~\citep{teng2022optimized}), and multi-stage frameworks (QME~\citep{qme2025}, FusionAgent~\citep{fusionagent2026}). See more details in the Appendix. 

\paragraph{Implementation details.}
Notably, our method requires no separate training split for fusion. Specifically, Ours (vanilla) performs purely identity-conditioned fusion, calibrating weights directly from the target gallery without prior training. Furthermore, following~\Secref{sec:method:composition}, we compose our framework with query-conditioned mechanisms from QME~\citep{qme2025} and FusionAgent~\citep{fusionagent2026}, reported as Ours (QME) and Ours (FusionAgent) respectively in the results, to demonstrate its plug-and-play compatibility. Across all datasets and model pools, we consistently fix hyperparameters to $\WeightSmoothing=0.5$, $\WeightTemperature=0.5$, and $\CompositionWeight=0.5$ in~\Eqref{eq:softmax} without dataset-specific tuning.

\subsection{Results}
\label{sec:exp:main}

\begin{table}[!t]
\caption{\textbf{Performance on CCVID and MEVID.} \emph{Comb.} specifies the model pool
shared by each block (\eFace~AdaFace, \eCAL~CAL, \eGait~BigGait on CCVID and AGRL on MEVID,
\eIns~InsightFace/ArcFace, and \eReIDL~LoRA-adapted CLIP-ReID); the top block reports individual-model results and the two below it the fusion methods. \best{Best} and \secondbest{second-best} fusion results
are marked within each pool; highlighted rows are ours. Values are in \%; FNIR is mean $\pm$
standard deviation over ten enrollment trials. R-1: Rank-1 accuracy, TAR: TAR@1\%FAR,
FNIR: FNIR@1\%FPIR.}
\label{tab:ccvid_mevid}
\begin{center}
\scriptsize
\tightrules
\setlength{\tabcolsep}{1.0pt}
\setlength{\combrowheight}{9.6pt}
\begin{subtable}[t]{0.49\linewidth}
\centering
\caption{CCVID}
\label{tab:ccvid}
\begin{tabular}{clcccc}
\toprule
\emph{Comb.} & Method & R-1 & mAP & TAR & FNIR ($\downarrow$) \\
\midrule
\comb{\eFace} & AdaFace & 94.0 & 87.9 & 75.7 & 13.0\pmm3.5 \\
\comb{\eCAL} & CAL & 81.4 & 74.7 & 66.3 & 52.8\pmm13.3 \\
\comb{\eGait} & BigGait & 76.7 & 61.0 & 49.7 & 71.1\pmm6.1 \\
\comb{\eIns} & InsightFace & 94.1 & 88.2 & 78.9 & 18.1\pmm3.1 \\
\midrule
 & QME & 94.1 & 90.8 & 76.2 & 12.3\pmm1.4 \\
 & FusionAgent & 93.4 & \secondbest{92.6} & \secondbest{85.9} & \secondbest{10.1\pmm1.5} \\
\ourrow & \ourslab{vanilla} & \best{95.3} & 92.1 & 81.0 & 11.0\pmm1.6 \\
\ourrow & \ourslab{QME} & \secondbest{95.2} & 91.9 & 79.3 & 10.3\pmm1.5 \\
\ourrow \combblock{5}{\eFace\eCAL\eGait} & \ourslab{FusionAgent} & 94.6 & \best{93.7} & \best{87.0} & \best{9.4\pmm1.2} \\
\midrule
 & Min-Fusion & 94.4 & 88.6 & 79.6 & 15.1\pmm2.6 \\
 & Max-Fusion & 84.2 & 69.2 & 57.5 & 51.2\pmm8.3 \\
 & Z-score & 93.5 & 90.8 & 80.9 & 13.2\pmm2.5 \\
 & Min-max & 93.4 & 90.9 & 78.8 & 13.5\pmm2.5 \\
 & RHE & 93.4 & 90.9 & 79.0 & 13.5\pmm2.3 \\
 & Asym-AO & 93.3 & 90.7 & 78.3 & 13.5\pmm2.5 \\
 & Farsight & 93.6 & 90.7 & 79.2 & 13.3\pmm2.0 \\
 & Weighted-sum & 94.2 & 89.8 & 81.5 & 13.5\pmm2.1 \\
 & BSSF & 93.5 & 90.6 & 82.8 & 15.0\pmm3.9 \\
 & QME & 93.0 & 91.5 & 78.9 & 11.2\pmm1.2 \\
\ourrow & \ourslab{vanilla} & \best{95.4} & \secondbest{92.2} & \best{86.0} & \best{9.2\pmm1.4} \\
\ourrow \combblock{12}{\eFace\eCAL\eGait\eIns} & \ourslab{QME} & \secondbest{94.7} & \best{92.7} & \secondbest{84.2} & \secondbest{9.6\pmm1.3} \\
\bottomrule
\end{tabular}
\end{subtable}
\hfill
\begin{subtable}[t]{0.49\linewidth}
\centering
\caption{MEVID}
\label{tab:mevid}
\begin{tabular}{clcccc}
\toprule
\emph{Comb.} & Method & R-1 & mAP & TAR & FNIR ($\downarrow$) \\
\midrule
\comb{\eFace} & AdaFace & 25.0 & 8.1 & 5.4 & 98.8\pmm1.2 \\
\comb{\eCAL} & CAL & 52.5 & 27.1 & 34.7 & 67.8\pmm7.3 \\
\comb{\eGait} & AGRL & 51.9 & 25.5 & 30.7 & 69.4\pmm8.9 \\
\comb{\eReIDL} & \shortstack[l]{CLIP-ReID} & 69.9 & 45.1 & 47.7 & 49.7\pmm4.3 \\
\midrule
 & BSSF & 54.4 & 28.7 & 33.4 & 63.8\pmm9.4 \\
 & QME & \secondbest{55.7} & 28.2 & 32.9 & 64.6\pmm8.2 \\
 & FusionAgent & 54.7 & 28.7 & 34.9 & \secondbest{58.6\pmm7.4} \\
\ourrow & \ourslab{vanilla} & 55.4 & \secondbest{31.0} & \secondbest{36.2} & 66.4\pmm13.6 \\
\ourrow & \ourslab{QME} & 54.7 & \secondbest{31.0} & 35.5 & 59.0\pmm12.6 \\
\ourrow \combblock{6}{\eFace\eCAL\eGait} & \ourslab{FusionAgent} & \best{57.0} & \best{31.2} & \best{37.3} & \best{52.0\pmm15.5} \\
\midrule
 & Min-Fusion & 47.8 & 21.7 & 28.0 & 70.4\pmm8.0 \\
 & Max-Fusion & \secondbest{68.4} & 43.9 & 45.9 & 55.8\pmm8.1 \\
 & Z-score & 66.8 & 41.1 & 44.9 & 52.1\pmm7.2 \\
 & Min-max & 66.1 & 40.3 & 42.6 & 54.8\pmm8.1 \\
 & RHE & 65.2 & 39.0 & 41.2 & 57.0\pmm8.7 \\
 & Asym-AO & 62.3 & 35.0 & 31.3 & 57.8\pmm8.0 \\
 & Farsight & 61.7 & 36.6 & 38.7 & 58.9\pmm9.1 \\
 & Weighted-sum & 65.5 & 40.4 & 45.2 & 52.6\pmm5.6 \\
 & QME & 65.5 & 42.1 & 46.3 & 51.2\pmm6.1 \\
\ourrow & \ourslab{vanilla} & \best{70.9} & \best{45.6} & \best{48.2} & \secondbest{49.0\pmm18.5} \\
\ourrow \combblock{11}{\eFace\eCAL\eGait\eReIDL} & \ourslab{QME} & 67.7 & \secondbest{44.2} & \secondbest{47.6} & \best{48.2\pmm16.8} \\
\bottomrule
\end{tabular}
\end{subtable}
\end{center}
\end{table}

Tables~\ref{tab:ccvid}, \ref{tab:mevid}, and~\ref{tab:ltcc} report recognition performance on CCVID~\citep{ccvid2022}, MEVID~\citep{mevid2023}, and LTCC~\citep{ltcc2020} across unimodal models, existing fusion baselines, and our framework. 
Overall, Ours (vanilla) consistently outperforms both strong single-modality baselines and supervised fusion schemes without using any training data. Specifically, on CCVID, our identity-conditioned-only fusion (vanilla) improves already strong backbones without additional training, pushing Rank-1 accuracy from $94.0\%$ (AdaFace) to $95.3\%$. Furthermore, composing with query-conditioned mechanisms consistently boosts performance across metrics: on MEVID, Ours (FusionAgent) improves upon standalone FusionAgent across all metrics, lifting mAP from $28.7\%$ to $31.2\%$ and TAR from $34.9\%$ to $37.3\%$. Also, our approach is particularly effective for open-set identification, reducing FNIR@$1\%$FPIR on LTCC from $60.7\%$ to $29.1\%$ via combined identity weighting and calibration.

\begin{figure}[t]
\centering
\includegraphics[width=\linewidth]{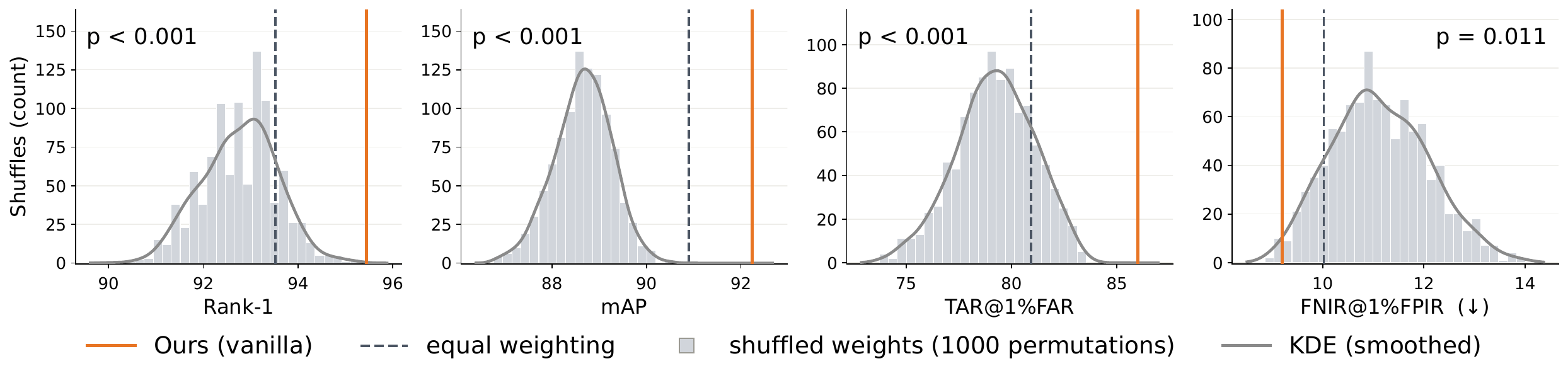}
\caption{\textbf{Identity profile permutation on CCVID.} Expert pool: AdaFace, CAL, BigGait, and InsightFace. Gray histograms show performance distributions over $1{,}000$ random identity-to-profile permutations while keeping models, gallery, and weight vectors identical. The genuine assignment (solid orange line) strictly outperforms all permutations across all three metrics, as well as equal weighting (dashed line), confirming that gains stem from identity-specific weights rather than arbitrary non-uniform weighting.}
\label{fig:permutation}
\end{figure}

\begin{table}[!t]
\begin{minipage}[t]{0.5\linewidth}
\caption{\textbf{Performance on LTCC.} \emph{Comb.} specifies the model pool shared by
each block (\eFace~AdaFace \citep{adaface2022}, \eCAL~CAL \citep{cal2022}, \eAIM~AIM
\citep{aim2023}, and \eReIDL~LoRA-adapted CLIP-ReID \citep{clipreid2023,lora2022}); the top
section reports individual-model results and the section below it the fusion methods.
\best{Best} and \secondbest{second-best} fusion results are marked; highlighted rows are ours.
Values are in \%; FNIR is mean $\pm$ standard deviation over ten enrollment trials.
TAR: TAR@1\%FAR, FNIR: FNIR@1\%FPIR.}
\label{tab:ltcc}
\begin{center}
\scriptsize
\setlength{\tabcolsep}{2pt}
\tightrules
\setlength{\combrowheight}{9.6pt}
\begin{tabular}{clcccc}
\toprule
\emph{Comb.} & Method & Rank-1 & mAP & TAR & FNIR ($\downarrow$) \\
\midrule
\comb{\eFace} & AdaFace & 18.5 & 5.9 & 2.4 & 99.8\pmm0.2 \\
\comb{\eCAL} & CAL & 74.4 & 40.6 & 36.7 & 59.7\pmm7.3 \\
\comb{\eAIM} & AIM & 74.8 & 40.9 & 37.0 & 66.2\pmm7.5 \\
\comb{\eReIDL} & CLIP-ReID (LoRA) & 76.7 & 40.7 & 31.9 & 42.0\pmm10.3 \\
\midrule
 & Min-Fusion & 42.0 & 15.0 & 13.1 & 81.9\pmm6.0 \\
 & Max-Fusion & 76.1 & 41.6 & 32.1 & 32.6\pmm11.0 \\
 & Z-score & 79.1 & 47.8 & 40.3 & 48.3\pmm13.3 \\
 & Min-max & 79.9 & 48.7 & 40.6 & 53.6\pmm15.3 \\
 & RHE & 78.3 & 46.3 & 38.4 & 64.9\pmm14.7 \\
 & Asym-AO & 76.3 & 38.8 & 32.2 & 63.0\pmm9.8 \\
 & Farsight & 80.3 & 48.6 & 40.4 & 49.9\pmm14.6 \\
 & Weighted-sum & 79.3 & 48.4 & \best{41.9} & 51.8\pmm9.1 \\
 & BSSF & 79.9 & \secondbest{49.3} & \best{41.9} & 50.9\pmm10.0 \\
 & QME & \best{80.7} & 48.5 & 41.3 & 60.7\pmm8.5 \\
\ourrow & \ourslab{vanilla} & \secondbest{80.5} & \best{49.6} & \secondbest{41.4} & \secondbest{31.1\pmm19.0} \\
\ourrow \combblock{12}{\eFace\eCAL\eAIM\eReIDL} & \ourslab{QME} & \secondbest{80.5} & 48.7 & 41.3 & \best{29.1\pmm21.2} \\
\bottomrule
\end{tabular}
\end{center}

\end{minipage}
\hfill
\begin{minipage}[t]{0.46\linewidth}
\caption{\textbf{Effectiveness of per-identity score calibration.} FNIR@$1\%$FPIR ($\downarrow$) of Ours (vanilla) before and after candidate calibration. Model symbols follow Tables~\ref{tab:ccvid_mevid} and~\ref{tab:ltcc}.}
\label{tab:calibration}
\begin{center}
\scriptsize
\setlength{\tabcolsep}{12pt}
\tightrules
\begin{tabular}{lcc}
\toprule
\emph{Comb.} & Before & After \\
\midrule
\multicolumn{3}{c}{\textit{Dataset: CCVID}} \\
\comb{\eFace\eCAL\eGait}        & 19.1\pmm8.1  & 11.0\pmm1.6 \\
\comb{\eFace\eCAL\eGait\eIns}   & 15.3\pmm7.3  & 9.2\pmm1.4 \\
\midrule
\multicolumn{3}{c}{\textit{Dataset: MEVID}} \\
\comb{\eFace\eCAL\eGait}        & 68.3\pmm14.3 & 66.4\pmm13.6 \\
\comb{\eFace\eCAL\eGait\eReIDL} & 52.6\pmm20.6 & 49.0\pmm18.5 \\
\midrule
\multicolumn{3}{c}{\textit{Dataset: LTCC}} \\
\comb{\eFace\eCAL\eAIM}         & 52.8\pmm12.5 & 52.0\pmm12.7 \\
\comb{\eFace\eCAL\eAIM\eReIDL}  & 37.3\pmm19.0 & 31.1\pmm19.0 \\
\bottomrule
\end{tabular}
\end{center}

\vspace{4pt}
\centering
\includegraphics[width=0.8\linewidth]{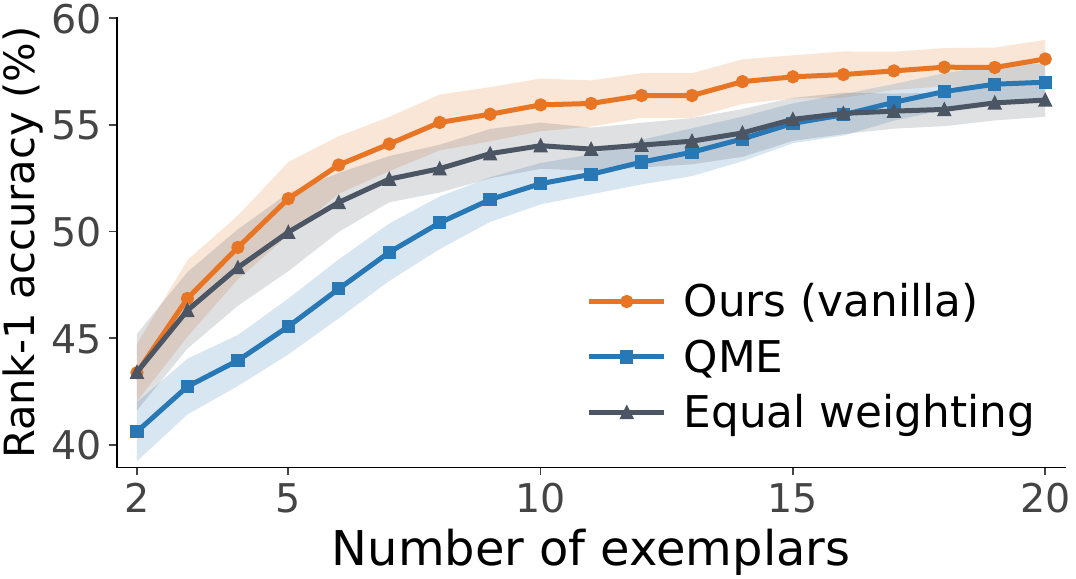}
\captionof{figure}{\textbf{Sensitivity to the number of exemplars.} Mean Rank-1 on MEVID (AdaFace, CAL, AGRL, InsightFace) over $30$ random exemplar draws; shading shows $\pm1$ SD. Ours (vanilla) leads at every exemplar count.}
\label{fig:exemplar_count}
\end{minipage}
\end{table}

\subsection{Ablation studies}
\label{sec:exp:ablations}

\paragraph{Effect of candidate score calibration.}
\label{sec:analysis:metrics}
To isolate the benefit of candidate calibration from identity-specific weighting, we evaluate open-set identification with and without this step. As shown in \Tabref{tab:calibration}, calibration consistently lowers FNIR@$1\%$FPIR across all configurations, cutting the error rate by up to $8.1\%$. On the CCVID dataset, for instance, it reduces FNIR from $19.1\%$ to $11.0\%$ while sharply compressing the cross-trial variance from $\pm8.1$ to $\pm1.6$. This confirms that while identity weighting provides the primary discriminative power, score calibration further improves non-mate rejection by stabilizing global thresholding across gallery identities.

\subsection{Analysis}
\label{sec:analysis:components}

\paragraph{Effectiveness of identity-specific weighting.}
To verify that the improvement of our method comes from tailoring the mixture profile to specific identities rather than just using non-uniform values, we perform a permutation test on CCVID. Specifically, we randomly shuffle the weight-to-identity assignments over $1{,}000$ independent runs so that each identity $\idIdx$ receives the weight vector estimated for a different identity $\auxIdIdx$, while keeping all models and gallery data identical. As shown in Figure~\ref{fig:permutation}, the matched assignment consistently outperforms both equal weighting and every random permutation across all three metrics. This confirms that the estimated weights encode meaningful, identity-specific priors rather than arbitrary non-uniform scaling.


\begin{figure}[t]
\centering
\includegraphics[width=\linewidth]{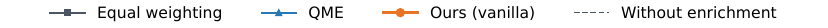}\\[2pt]
\begin{subfigure}[b]{0.42\linewidth}\centering\includegraphics[width=\linewidth]{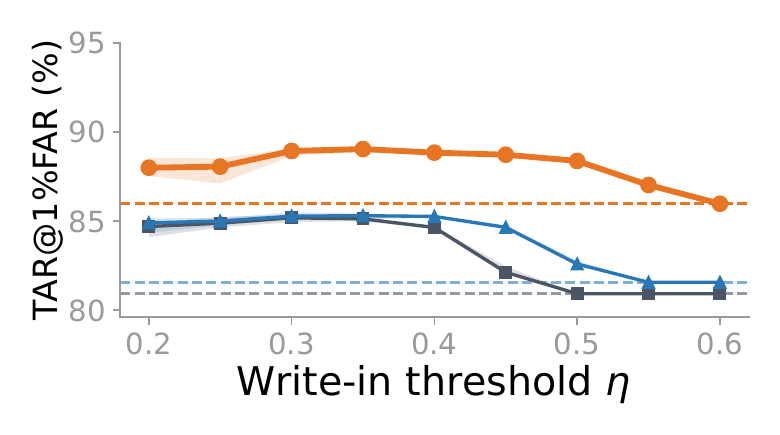}\caption{}\end{subfigure}
\begin{subfigure}[b]{0.42\linewidth}\centering\includegraphics[width=\linewidth]{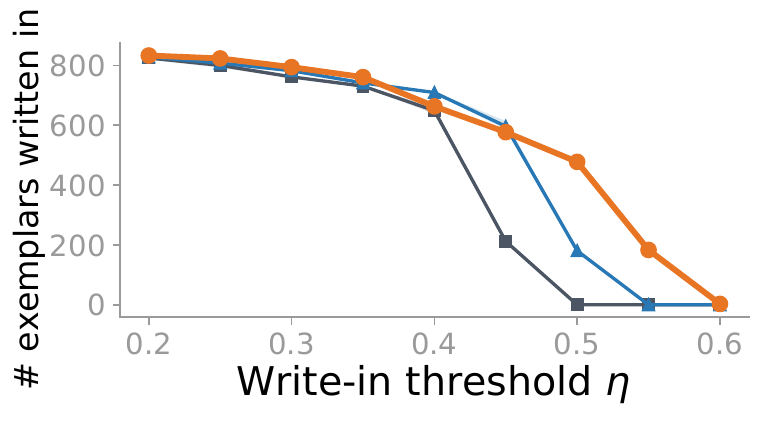}\caption{}\end{subfigure}
\caption{\textbf{Robustness to gallery enrichment.} Queries are added into the gallery under their predicted identity $\hat{\idIdx}$ when the calibrated match score satisfies $S_{\query,\hat{\idIdx}} > \eta$. Solid curves and shaded regions denote the mean and spread across 9 arrival sequences, while dashed lines mark the static-gallery baselines. We report: (a) TAR@$1\%$FAR and (b) number of admitted queries across enrollment thresholds $\eta$. Ours (vanilla) consistently outperforms both equal weighting and QME across all thresholds, remaining robust regardless of enrichment aggressiveness.}

\label{fig:enriched}
\end{figure}

\paragraph{Sensitivity to the number of exemplars per identity.}
Because identity-specific weights are estimated directly from gallery comparisons, we evaluate whether our method relies on having densely populated galleries. On a subset of MEVID ($36$ identities with sufficient samples, evaluated across $251$ queries), we vary the number of enrolled exemplars per identity from $2$ to $20$ over $30$ paired random draws. As shown in Figure~\ref{fig:exemplar_count}, Ours (vanilla) consistently outperforms both equal weighting and QME across all exemplar counts. While overall performance scales with more exemplars for all methods, our advantage remains steady across all everage exemplar counts. This confirms that identity-specific weighting does not require extensive reference sets to be reliable, exhibiting strong robustness even in data-scarce enrollment regimes.

\paragraph{Robustness to gallery enrichment.}
\label{sec:analysis:enriched}
Standard evaluation protocol assumes a static gallery where queries are discarded after scoring. To test whether identity-specific weighting remains robust when the gallery evolves dynamically, we evaluate recognition under gallery enrichment on CCVID using AdaFace, CAL, BigGait, and InsightFace. In each evaluation stream, a query is admitted as a new gallery exemplar if its calibrated top score exceeds an enrollment threshold $\eta$, \ie when $\idScore{\query}{\matchedIdx} > \eta$ after evaluatin. For Ours (vanilla), newly admitted exemplars immediately update the identity's fusion weights without retraining. Because incoming queries can introduce pseudo-label noise and dynamic updates depend on arrival sequence, we evaluate across 9 random arrival orders and report both the mean and range. 
As shown in \Figref{fig:enriched}, all methods benefit from gallery enrichment at low-to-moderate thresholds ($\eta \le 4$) before converging to their static-gallery baselines as $\eta$ increases and fewer queries qualify for admission. Notably, Ours (vanilla) consistently maintains a wide margin over equal weighting and QME across the entire threshold range, remaining resilient even when low thresholds introduce potential misidentifications. Experiments under an alternative write-in criterion show consistent gains and are reported in the Appendix.
\section{Conclusion}
\label{sec:conclusion}
In this work, we demonstrated that multi-model fusion for open-set identification substantially benefits from identity-specific reliability priors estimated directly from the given gallery exemplars. By exploiting intra- and inter-identity separation statistics within the gallery, our approach tailors model weights to individual identities without requiring additional training. Across multiple model pools and challenging clothes-changing benchmarks, our identity-specific weighting consistently outperforms uniform and quality-based baselines, seamlessly integrates with query-dependent fusion, and exhibits strong robustness under label noise, data scarcity, and dynamic gallery evolution. 

\paragraph{Limitations.}
Our method estimates each model's reliability profile individually based on gallery comparisons, without explicitly modeling cross-model feature correlations or statistical redundancies among the candidate backbones. Extending this lightweight estimation to capture joint multi-model dependencies presents a promising direction for further refinement. Additionally, while candidate score calibration effectively aligns open-set similarity scales across models, open-set verification at ultra-low false acceptance regimes remains constrained by the limited non-mate query scale in existing clothes-changing benchmarks.

\bibliography{iclr2027_conference}

@inproceedings{adaface2022,
  title     = {AdaFace: Quality Adaptive Margin for Face Recognition},
  author    = {Kim, Minchul and Jain, Anil K. and Liu, Xiaoming},
  booktitle = {Proceedings of the IEEE/CVF Conference on Computer Vision and Pattern Recognition (CVPR)},
  pages     = {18750--18759},
  year      = {2022},
  doi       = {10.1109/CVPR52688.2022.01819}
}

@inproceedings{cal2022,
  title     = {Clothes-Changing Person Re-identification with RGB Modality Only},
  author    = {Gu, Xinqian and Chang, Hong and Ma, Bingpeng and Bai, Shutao and Shan, Shiguang and Chen, Xilin},
  booktitle = {Proceedings of the IEEE/CVF Conference on Computer Vision and Pattern Recognition (CVPR)},
  pages     = {1060--1069},
  year      = {2022},
  doi       = {10.1109/CVPR52688.2022.00113}
}

@inproceedings{ccvid2022,
  title     = {Clothes-Changing Person Re-identification with RGB Modality Only},
  author    = {Gu, Xinqian and Chang, Hong and Ma, Bingpeng and Bai, Shutao and Shan, Shiguang and Chen, Xilin},
  booktitle = {Proceedings of the IEEE/CVF Conference on Computer Vision and Pattern Recognition (CVPR)},
  pages     = {1060--1069},
  year      = {2022},
  doi       = {10.1109/CVPR52688.2022.00113}
}

@inproceedings{mevid2023,
  title     = {MEVID: Multi-view Extended Videos with Identities for Video Person Re-Identification},
  author    = {Davila, Daniel and Du, Dawei and Lewis, Bryon and Funk, Christopher and Van Pelt, Joseph and Collins, Roderic and Corona, Kellie and Brown, Matthew and McCloskey, Scott and Hoogs, Anthony and Clipp, Brian},
  booktitle = {Proceedings of the IEEE/CVF Winter Conference on Applications of Computer Vision (WACV)},
  pages     = {1634--1643},
  year      = {2023},
  doi       = {10.1109/WACV56688.2023.00168}
}

@inproceedings{biggait2024,
  title     = {BigGait: Learning Gait Representation You Want by Large Vision Models},
  author    = {Ye, Dingqiang and Fan, Chao and Ma, Jingzhe and Liu, Xiaoming and Yu, Shiqi},
  booktitle = {Proceedings of the IEEE/CVF Conference on Computer Vision and Pattern Recognition (CVPR)},
  pages     = {200--210},
  year      = {2024},
  doi       = {10.1109/CVPR52733.2024.00027}
}

@inproceedings{clipreid2023,
  title     = {CLIP-ReID: Exploiting Vision-Language Model for Image Re-Identification without Concrete Text Labels},
  author    = {Li, Siyuan and Sun, Li and Li, Qingli},
  booktitle = {Proceedings of the AAAI Conference on Artificial Intelligence},
  volume    = {37},
  pages     = {1405--1413},
  year      = {2023},
  doi       = {10.1609/aaai.v37i1.25225}
}

@inproceedings{wei2018ptgan,
  title     = {Person Transfer GAN to Bridge Domain Gap for Person Re-Identification},
  author    = {Wei, Longhui and Zhang, Shiliang and Gao, Wen and Tian, Qi},
  booktitle = {Proceedings of the IEEE/CVF Conference on Computer Vision and Pattern Recognition (CVPR)},
  pages     = {79--88},
  year      = {2018}
}

@inproceedings{zheng2015market,
  title     = {Scalable Person Re-Identification: A Benchmark},
  author    = {Zheng, Liang and Shen, Liyue and Tian, Lu and Wang, Shengjin and Wang, Jingdong and Tian, Qi},
  booktitle = {Proceedings of the IEEE International Conference on Computer Vision (ICCV)},
  pages     = {1116--1124},
  year      = {2015}
}

@inproceedings{gou2017dukemtmc,
  title     = {{DukeMTMC4ReID}: A Large-Scale Multi-Camera Person Re-Identification Dataset},
  author    = {Gou, Mengran and Karanam, Srikrishna and Liu, Wenqian and Camps, Octavia and Radke, Richard J.},
  booktitle = {Proceedings of the IEEE Conference on Computer Vision and Pattern Recognition Workshops},
  year      = {2017}
}

@inproceedings{lora2022,
  title         = {LoRA: Low-Rank Adaptation of Large Language Models},
  author        = {Hu, Edward J. and Shen, Yelong and Wallis, Phillip and Allen-Zhu, Zeyuan and Li, Yuanzhi and Wang, Shean and Wang, Lu and Chen, Weizhu},
  booktitle     = {International Conference on Learning Representations (ICLR)},
  year          = {2022},
  eprint        = {2106.09685},
  archivePrefix = {arXiv}
}

@InProceedings{qme2025,
    author    = {Zhu, Jie and Su, Yiyang and Kim, Minchul and Jain, Anil and Liu, Xiaoming},
    title     = {A Quality-Guided Mixture of Score-Fusion Experts Framework for Human Recognition},
    booktitle = {Proceedings of the IEEE/CVF International Conference on Computer Vision (ICCV)},
    month     = {October},
    year      = {2025},
    pages     = {13076-13086},
    url       = {https://openaccess.thecvf.com/content/ICCV2025/html/Zhu_A_Quality-Guided_Mixture_of_Score-Fusion_Experts_Framework_for_Human_Recognition_ICCV_2025_paper.html},
    doi = {10.1109/ICCV51701.2025.01215},
    eprint = {2508.00053},
    archivePrefix = {arXiv}
}

@inproceedings{fusionagent2026,
  title         = {FusionAgent: A Multimodal Agent with Dynamic Model Selection for Human Recognition},
  author        = {Zhu, Jie and Guo, Xiao and Su, Yiyang and Jain, Anil K. and Liu, Xiaoming},
  booktitle     = {Proceedings of the IEEE/CVF Conference on Computer Vision and Pattern Recognition (CVPR)},
  year          = {2026},
  eprint        = {2603.26908},
  archivePrefix = {arXiv}
}

@article{agrl,
  title   = {Adaptive Graph Representation Learning for Video Person Re-Identification},
  author  = {Wu, Yiming and Bourahla, Omar El Farouk and Li, Xi and Wu, Fei and Tian, Qi and Zhou, Xue},
  journal = {IEEE Transactions on Image Processing},
  volume  = {29},
  pages   = {8821--8830},
  year    = {2020},
  doi     = {10.1109/TIP.2020.3001693}
}

@article{jain2005score,
  title   = {Score normalization in multimodal biometric systems},
  author  = {Jain, Anil K. and Nandakumar, Karthik and Ross, Arun},
  journal = {Pattern Recognition},
  volume  = {38},
  number  = {12},
  pages   = {2270--2285},
  year    = {2005},
  doi     = {10.1016/j.patcog.2005.01.012}
}

@article{auckenthaler2000score,
  title   = {Score Normalization for Text-Independent Speaker Verification Systems},
  author  = {Auckenthaler, Roland and Carey, Michael and Lloyd-Thomas, Harvey},
  journal = {Digital Signal Processing},
  volume  = {10},
  number  = {1--3},
  pages   = {42--54},
  year    = {2000},
  doi     = {10.1006/dspr.1999.0360}
}

@inproceedings{park2021weighted,
  title     = {Multi-scale speaker diarization with neural affinity score fusion},
  author    = {Park, Tae Jin and Kumar, Manoj and Narayanan, Shrikanth},
  booktitle = {IEEE International Conference on Acoustics, Speech and Signal Processing (ICASSP)},
  year      = {2021}
}

@article{teng2022optimized,
  title   = {Optimized Score Level Fusion for Multi-Instance Finger Vein Recognition},
  author  = {Teng, J. H. and Ong, T. S. and Connie, T. and Sonai Muthu Anbananthen, K. and Min, P. P.},
  journal = {Algorithms},
  volume  = {15},
  number  = {5},
  pages   = {161},
  year    = {2022},
  doi     = {10.3390/a15050161}
}

@inproceedings{snelick2003multimodal,
  title     = {Multimodal biometrics: issues in design and testing},
  author    = {Snelick, Robert and Indovina, Mike and Yen, James and Mink, Alan},
  booktitle = {International Conference on Multimodal Interfaces (ICMI)},
  year      = {2003}
}

@article{he2010performance,
  title   = {Performance evaluation of score level fusion in multimodal biometric systems},
  author  = {He, Mingxing and Horng, Shi-Jinn and Fan, Pingzhi and Run, Ray-Shine and Chen, Rong-Jian and Lai, Jui-Lin and Khan, Muhammad Khurram and Sentosa, Kevin Octavius},
  journal = {Pattern Recognition},
  volume  = {43},
  number  = {5},
  pages   = {1789--1800},
  year    = {2010}
}

@article{herbadji2020combining,
  title   = {Combining multiple biometric traits using asymmetric aggregation operators for improved person recognition},
  author  = {Herbadji, Abderrahmane and Akhtar, Zahid and Siddique, Kamran and Guermat, Noubeil and Ziet, Lahcene and Cheniti, Mohamed and Muhammad, Khan},
  journal = {Symmetry},
  volume  = {12},
  number  = {3},
  pages   = {444},
  year    = {2020}
}

@misc{liu2025person,
      title={Person Recognition at Altitude and Range: Fusion of Face, Body Shape and Gait}, 
      author={Feng Liu and Nicholas Chimitt and Lanqing Guo and Jitesh Jain and Aditya Kane and Minchul Kim and Wes Robbins and Yiyang Su and Dingqiang Ye and Xingguang Zhang and Jie Zhu and Siddharth Satyakam and Christopher Perry and Stanley H. Chan and Arun Ross and Humphrey Shi and Zhangyang Wang and Anil Jain and Xiaoming Liu},
      year={2025},
      eprint={2505.04616},
      archivePrefix={arXiv},
      primaryClass={cs.CV},
      url={https://arxiv.org/abs/2505.04616}, 
}

@inproceedings{doddington1998sheep,
  title     = {{SHEEP}, {GOATS}, {LAMBS} and {WOLVES}: A Statistical Analysis of Speaker
               Performance in the {NIST} 1998 Speaker Recognition Evaluation},
  author    = {Doddington, George and Liggett, Walter and Martin, Alvin and
               Przybocki, Mark and Reynolds, Douglas A.},
  booktitle = {5th International Conference on Spoken Language Processing (ICSLP)},
  year      = {1998},
  doi       = {10.21437/ICSLP.1998-244}
}

@article{yager2010menagerie,
  title   = {The Biometric Menagerie},
  author  = {Yager, Neil and Dunstone, Ted},
  journal = {IEEE Transactions on Pattern Analysis and Machine Intelligence},
  volume  = {32},
  number  = {2},
  pages   = {220--230},
  year    = {2010},
  doi     = {10.1109/TPAMI.2008.291}
}

@article{zheng2016openworld,
  title   = {Towards Open-World Person Re-Identification by One-Shot Group-Based
             Verification},
  author  = {Zheng, Wei-Shi and Gong, Shaogang and Xiang, Tao},
  journal = {IEEE Transactions on Pattern Analysis and Machine Intelligence},
  volume  = {38},
  number  = {3},
  pages   = {591--606},
  year    = {2016},
  doi     = {10.1109/TPAMI.2015.2453984}
}

@techreport{grother2019frvt,
  title       = {Face Recognition Vendor Test ({FRVT}) Part 2: Identification},
  author      = {Grother, Patrick and Ngan, Mei and Hanaoka, Kayee},
  institution = {National Institute of Standards and Technology},
  number      = {NISTIR 8271},
  year        = {2019},
  doi         = {10.6028/NIST.IR.8271}
}

@inproceedings{ltcc2020,
  title     = {Long-Term Cloth-Changing Person Re-identification},
  author    = {Qian, Xuelin and Wang, Wenxuan and Zhang, Li and Zhu, Fangrui and
               Fu, Yanwei and Xiang, Tao and Jiang, Yu-Gang and Xue, Xiangyang},
  booktitle = {Proceedings of the Asian Conference on Computer Vision (ACCV)},
  pages     = {71--88},
  year      = {2020},
  doi       = {10.1007/978-3-030-69535-4_5}
}

@article{prcc2021,
  title   = {Person Re-identification by Contour Sketch under Moderate Clothing Change},
  author  = {Yang, Qize and Wu, Ancong and Zheng, Wei-Shi},
  journal = {IEEE Transactions on Pattern Analysis and Machine Intelligence},
  volume  = {43},
  number  = {6},
  pages   = {2029--2046},
  year    = {2021},
  doi     = {10.1109/TPAMI.2019.2960509}
}

@inproceedings{aim2023,
  title     = {Good is Bad: Causality Inspired Cloth-Debiasing for Cloth-Changing
               Person Re-identification},
  author    = {Yang, Zhengwei and Lin, Meng and Zhong, Xian and Wu, Yu and Wang, Zheng},
  booktitle = {Proceedings of the IEEE/CVF Conference on Computer Vision and Pattern
               Recognition (CVPR)},
  pages     = {1472--1481},
  year      = {2023}
}

@inproceedings{arcface2019,
  title     = {ArcFace: Additive Angular Margin Loss for Deep Face Recognition},
  author    = {Deng, Jiankang and Guo, Jia and Xue, Niannan and Zafeiriou, Stefanos},
  booktitle = {Proceedings of the IEEE/CVF Conference on Computer Vision and Pattern
               Recognition (CVPR)},
  pages     = {4690--4699},
  year      = {2019},
  doi       = {10.1109/CVPR.2019.00482}
}

@article{gupta2020quality,
  title   = {Quality based adaptive score fusion approach for multimodal biometric system},
  author  = {Gupta, Keshav and Walia, Gurjit Singh and Sharma, Kapil},
  journal = {Applied Intelligence},
  volume  = {50},
  number  = {4},
  pages   = {1086--1099},
  year    = {2020},
  doi     = {10.1007/s10489-019-01579-1}
}

@article{poh2009benchmarking,
  title   = {Benchmarking Quality-Dependent and Cost-Sensitive Score-Level Multimodal Biometric Fusion Algorithms},
  author  = {Poh, Norman and Bourlai, Thirimachos and Kittler, Josef and Allano, Lorene and Alonso-Fernandez, Fernando and Ambekar, Onkar and Baker, John and Dorizzi, Bernadette and Fatukasi, Omolara and Fierrez, Julian and Ganster, Harald and Ortega-Garcia, Javier and Maurer, Donald and Salah, Albert Ali and Scheidat, Tobias and Vielhauer, Claus},
  journal = {IEEE Transactions on Information Forensics and Security},
  volume  = {4},
  number  = {4},
  pages   = {849--866},
  year    = {2009},
  doi     = {10.1109/TIFS.2009.2034885}
}

@article{fierrez2005bayesian,
  title   = {Bayesian adaptation for user-dependent multimodal biometric authentication},
  author  = {Fierrez-Aguilar, Julian and Garcia-Romero, Daniel and Ortega-Garcia, Javier and Gonzalez-Rodriguez, Joaquin},
  journal = {Pattern Recognition},
  volume  = {38},
  number  = {8},
  pages   = {1317--1319},
  year    = {2005},
  doi     = {10.1016/j.patcog.2005.01.013}
}

@inproceedings{hamobe2025,
  title         = {{HAMoBE}: Hierarchical and Adaptive Mixture of Biometric Experts for Video-based Person {ReID}},
  author        = {Su, Yiyang and Shi, Yunping and Liu, Feng and Liu, Xiaoming},
  booktitle     = {Proceedings of the IEEE/CVF International Conference on Computer Vision (ICCV)},
  pages         = {11525--11536},
  year          = {2025},
  eprint        = {2508.05038},
  archivePrefix = {arXiv}
}

@book{green1966signal,
  title     = {Signal Detection Theory and Psychophysics},
  author    = {Green, David M. and Swets, John A.},
  publisher = {John Wiley \& Sons},
  address   = {New York},
  year      = {1966}
}

@article{snelick2005user, title={Large-scale evaluation of multimodal biometric authentication using state-of-the-art systems}, volume={27}, ISSN={0162-8828}, url={http://dx.doi.org/10.1109/TPAMI.2005.57}, DOI={10.1109/tpami.2005.57}, number={3}, journal={IEEE Transactions on Pattern Analysis and Machine Intelligence}, publisher={Institute of Electrical and Electronics Engineers (IEEE)}, author={Snelick, R. and Uludag, U. and Mink, A. and Indovina, M. and Jain, A.}, year={2005}, month=Mar, pages={450--455} }

@article{poh2013user, title={A user-specific and selective multimodal biometric fusion strategy by ranking subjects}, volume={46}, ISSN={0031-3203}, url={http://dx.doi.org/10.1016/j.patcog.2013.03.018}, DOI={10.1016/j.patcog.2013.03.018}, number={12}, journal={Pattern Recognition}, publisher={Elsevier BV}, author={Poh, Norman and Ross, Arun and Lee, Weifeng and Kittler, Josef}, year={2013}, month=Dec, pages={3341--3357} }

@InProceedings{invarreid2023,
    author    = {Liu, Feng and Kim, Minchul and Gu, ZiAng and Jain, Anil and Liu, Xiaoming},
    title     = {{Learning Clothing and Pose Invariant 3D Shape Representation for Long-Term Person Re-Identification}},
    booktitle = {Proceedings of the IEEE/CVF International Conference on Computer Vision (ICCV)},
    month     = {October},
    year      = {2023},
    pages     = {19617-19626}
}

@InProceedings{clip3dreid2024,
    author    = {Liu, Feng and Kim, Minchul and Ren, Zhiyuan and Liu, Xiaoming},
    title     = {{Distilling CLIP with Dual Guidance for Learning Discriminative Human Body Shape Representation}},
    booktitle = {Proceedings of the IEEE/CVF Conference on Computer Vision and Pattern Recognition (CVPR)},
    month     = {June},
    year      = {2024},
    pages     = {256-266}
}

@InProceedings{differ2025,
    author    = {Liang, Xin and Rawat, Yogesh S},
    title     = {{DIFFER: Disentangling Identity Features via Semantic Cues for Clothes-Changing Person Re-ID}},
    booktitle = {Proceedings of the IEEE/CVF Conference on Computer Vision and Pattern Recognition (CVPR)},
    month     = {June},
    year      = {2025},
    pages     = {13980-13989}
}

@InProceedings{csci2025,
    author    = {Pathak, Priyank and Rawat, Yogesh S.},
    title     = {{Colors See Colors Ignore: Clothes Changing ReID with Color Disentanglement}},
    booktitle = {Proceedings of the IEEE/CVF International Conference on Computer Vision (ICCV)},
    month     = {October},
    year      = {2025},
    pages     = {16797-16807},
    url       = {https://openaccess.thecvf.com/content/ICCV2025/html/Pathak_Colors_See_Colors_Ignore_Clothes_Changing_ReID_with_Color_Disentanglement_ICCV_2025_paper.html}
}

@InProceedings{ccpg2023,
    author    = {Li, Weijia and Hou, Saihui and Zhang, Chunjie and Cao, Chunshui and Liu, Xu and Huang, Yongzhen and Zhao, Yao},
    title     = {{An In-Depth Exploration of Person Re-Identification and Gait Recognition in Cloth-Changing Conditions}},
    booktitle = {Proceedings of the IEEE/CVF Conference on Computer Vision and Pattern Recognition (CVPR)},
    month     = {June},
    year      = {2023},
    pages     = {13824-13833}
}

@InProceedings{farsight2024,
    author    = {Liu, Feng and Ashbaugh, Ryan and Chimitt, Nicholas and Hassan, Najmul and Hassani, Ali and Jaiswal, Ajay and Kim, Minchul and Mao, Zhiyuan and Perry, Christopher and Ren, Zhiyuan and Su, Yiyang and Varghaei, Pegah and Wang, Kai and Zhang, Xingguang and Chan, Stanley and Ross, Arun and Shi, Humphrey and Wang, Zhangyang and Jain, Anil and Liu, Xiaoming},
    title     = {{FarSight: A Physics-Driven Whole-Body Biometric System at Large Distance and Altitude}},
    booktitle = {Proceedings of the IEEE/CVF Winter Conference on Applications of Computer Vision (WACV)},
    month     = {January},
    year      = {2024},
    pages     = {6227-6236}
}

@InProceedings{sapiensid2025,
    author    = {Kim, Minchul and Ye, Dingqiang and Su, Yiyang and Liu, Feng and Liu, Xiaoming},
    title     = {{SapiensID: Foundation for Human Recognition}},
    booktitle = {Proceedings of the IEEE/CVF Conference on Computer Vision and Pattern Recognition (CVPR)},
    month     = {June},
    year      = {2025},
    pages     = {13937-13947},
    url       = {https://openaccess.thecvf.com/content/CVPR2025/html/Kim_SapiensID_Foundation_for_Human_Recognition_CVPR_2025_paper.html}
}

@inproceedings{openset2024, title={{Open-Set Biometrics: Beyond Good Closed-Set Models}}, ISBN={9783031730337}, ISSN={1611-3349}, url={http://dx.doi.org/10.1007/978-3-031-73033-7_14}, DOI={10.1007/978-3-031-73033-7_14}, booktitle={Computer Vision -- ECCV 2024}, publisher={Springer Nature Switzerland}, author={Su, Yiyang and Kim, Minchul and Liu, Feng and Jain, Anil and Liu, Xiaoming}, year={2024}, month=Oct, pages={243--261} }

@misc{localscore2026,
      title={{LocalScore: Local Density-Aware Similarity Scoring for Biometrics}},
      author={Yiyang Su and Minchul Kim and Jie Zhu and Christopher Perry and Feng Liu and Anil Jain and Xiaoming Liu},
      year={2026},
      eprint={2602.01012},
      archivePrefix={arXiv},
      primaryClass={cs.CV},
      url={https://arxiv.org/abs/2602.01012},
}

@misc{liao2014openset,
      title={Open-set Person Re-identification}, 
      author={Shengcai Liao and Zhipeng Mo and Jianqing Zhu and Yang Hu and Stan Z. Li},
      year={2014},
      eprint={1408.0872},
      archivePrefix={arXiv},
      primaryClass={cs.CV},
      url={https://arxiv.org/abs/1408.0872}, 
}

@article{ross2003fusion, title={Information fusion in biometrics}, volume={24}, ISSN={0167-8655}, url={http://dx.doi.org/10.1016/S0167-8655(03)00079-5}, DOI={10.1016/s0167-8655(03)00079-5}, number={13}, journal={Pattern Recognition Letters}, publisher={Elsevier BV}, author={Ross, Arun and Jain, Anil}, year={2003}, month=sep, pages={2115--2125} }

@article{nandakumar2008likelihood, title={Likelihood Ratio-Based Biometric Score Fusion}, volume={30}, ISSN={0162-8828}, url={http://dx.doi.org/10.1109/TPAMI.2007.70796}, DOI={10.1109/tpami.2007.70796}, number={2}, journal={IEEE Transactions on Pattern Analysis and Machine Intelligence}, publisher={Institute of Electrical and Electronics Engineers (IEEE)}, author={Nandakumar, K. and Yi Chen and Dass, S.C. and Jain, A.K.}, year={2008}, month=Feb, pages={342--347} }

@inproceedings{nandakumar2006quality, title={Quality-based Score Level Fusion in Multibiometric Systems}, url={http://dx.doi.org/10.1109/ICPR.2006.951}, DOI={10.1109/icpr.2006.951}, booktitle={18th International Conference on Pattern Recognition (ICPR'06)}, publisher={IEEE}, author={Nandakumar, K. and Yi Chen and Jain, A.K. and Dass, S.C.}, year={2006} }

@article{poh2012unified, title={A Unified Framework for Biometric Expert Fusion Incorporating Quality Measures}, volume={34}, ISSN={2160-9292}, url={http://dx.doi.org/10.1109/TPAMI.2011.102}, DOI={10.1109/tpami.2011.102}, number={1}, journal={IEEE Transactions on Pattern Analysis and Machine Intelligence}, publisher={Institute of Electrical and Electronics Engineers (IEEE)}, author={Poh, N. and Kittler, J.}, year={2012}, month=Jan, pages={3--18} }

@InProceedings{seas2024,
    author    = {Zhu, Haidong and Budhwant, Pranav and Zheng, Zhaoheng and Nevatia, Ram},
    title     = {{SEAS: ShapE-Aligned Supervision for Person Re-Identification}},
    booktitle = {Proceedings of the IEEE/CVF Conference on Computer Vision and Pattern Recognition (CVPR)},
    month     = {June},
    year      = {2024},
    pages     = {164--174},
    url       = {https://openaccess.thecvf.com/content/CVPR2024/html/Zhu_SEAS_ShapE-Aligned_Supervision_for_Person_Re-Identification_CVPR_2024_paper.html}
}

@InProceedings{icsm2025,
    author    = {Pang, Zhiqi and Wang, Junjie and Zhao, Lingling and Wang, Chunyu},
    title     = {{Identity-Clothing Similarity Modeling for Unsupervised Clothing Change Person Re-Identification}},
    booktitle = {Proceedings of the IEEE/CVF Conference on Computer Vision and Pattern Recognition (CVPR)},
    month     = {June},
    year      = {2025},
    pages     = {19251--19260},
    url       = {https://openaccess.thecvf.com/content/CVPR2025/html/Pang_Identity-Clothing_Similarity_Modeling_for_Unsupervised_Clothing_Change_Person_Re-Identification_CVPR_2025_paper.html}
}

@InProceedings{gaitmax2026,
    author    = {Huang, Zhanbo and Ye, Dingqiang and Liu, Xiaoming and Kong, Yu},
    title     = {{Unlocking Motion from Large Vision Models with a Semantic and Kinematic Duality for Gait Recognition}},
    booktitle = {Proceedings of the IEEE/CVF Conference on Computer Vision and Pattern Recognition (CVPR)},
    month     = {June},
    year      = {2026},
    pages     = {28379--28390},
    url       = {https://openaccess.thecvf.com/content/CVPR2026/html/Huang_Unlocking_Motion_from_Large_Vision_Models_with_a_Semantic_and_CVPR_2026_paper.html}
}

@InProceedings{pdf2024,
  title = 	 {{Predictive Dynamic Fusion}},
  author =       {Cao, Bing and Xia, Yinan and Ding, Yi and Zhang, Changqing and Hu, Qinghua},
  booktitle = 	 {Proceedings of the 41st International Conference on Machine Learning},
  pages = 	 {5608--5628},
  year = 	 {2024},
  editor = 	 {Salakhutdinov, Ruslan and Kolter, Zico and Heller, Katherine and Weller, Adrian and Oliver, Nuria and Scarlett, Jonathan and Berkenkamp, Felix},
  volume = 	 {235},
  series = 	 {Proceedings of Machine Learning Research},
  month = 	 {21--27 Jul},
  publisher =    {PMLR},
  url = 	 {https://proceedings.mlr.press/v235/cao24c.html},
  
}

@inproceedings{cocoon2025,
 author = {Cho, Minkyoung and Cao, Yulong and Sun, Jiachen and Zhang, Qingzhao and Pavone, Marco and Park, Jeong Joon and Yang, Heng and Mao, Zhuoqing},
 booktitle = {International Conference on Learning Representations},
 editor = {Y. Yue and A. Garg and N. Peng and F. Sha and R. Yu},
 pages = {40998--41021},
 title = {{Cocoon: Robust Multi-Modal Perception with Uncertainty-Aware Sensor Fusion}},
 url = {https://proceedings.iclr.cc/paper_files/paper/2025/file/6580845ce91a748a9bb09cec079aac1d-Paper-Conference.pdf},
 volume = {2025},
 year = {2025}
}

@InProceedings{evidencefilter2025,
    author    = {Liu, Wei and Chen, Yufei and Yue, Xiaodong},
    title     = {{Enhancing Testing-Time Robustness for Trusted Multi-View Classification in the Wild}},
    booktitle = {Proceedings of the IEEE/CVF Conference on Computer Vision and Pattern Recognition (CVPR)},
    month     = {June},
    year      = {2025},
    pages     = {15508--15517},
    url       = {https://openaccess.thecvf.com/content/CVPR2025/html/Liu_Enhancing_Testing-Time_Robustness_for_Trusted_Multi-View_Classification_in_the_Wild_CVPR_2025_paper.html}
}

@ARTICLE{scheirer2013toward,
  author={Scheirer, Walter J. and de Rezende Rocha, Anderson and Sapkota, Archana and Boult, Terrance E.},
  journal={IEEE Transactions on Pattern Analysis and Machine Intelligence}, 
  title={Toward Open Set Recognition}, 
  year={2013},
  volume={35},
  number={7},
  pages={1757-1772},
  doi={10.1109/TPAMI.2012.256}}

@misc{bendale2016towards,
      title={Towards Open Set Deep Networks}, 
      author={Abhijit Bendale and Terrance Boult},
      year={2015},
      eprint={1511.06233},
      archivePrefix={arXiv},
      primaryClass={cs.CV},
      url={https://arxiv.org/abs/1511.06233}, 
}

@inproceedings{dosovitskiy2021vit,
  author = {Dosovitskiy, Alexey and Beyer, Lucas and Kolesnikov, Alexander and Weissenborn, Dirk and Zhai, Xiaohua and Unterthiner, Thomas and Dehghani, Mostafa and Minderer, Matthias and Heigold, Georg and Gelly, Sylvain and Uszkoreit, Jakob and Houlsby, Neil},
  title = {An Image is Worth 16x16 Words: Transformers for Image Recognition at Scale},
  booktitle = {International Conference on Learning Representations},
  year = {2021},
  url = {https://arxiv.org/abs/2010.11929}
}

@inproceedings{zhu2021webface260m,
  author = {Zhu, Zheng and Huang, Guan and Deng, Jiankang and Ye, Yun and Huang, Junjie and Chen, Xinze and Zhu, Jiagang and Yang, Tian and Lu, Jiwen and Du, Dalong and Zhou, Jie},
  title = {{WebFace260M}: A Benchmark Unveiling the Power of Million-Scale Deep Face Recognition},
  booktitle = {Proceedings of the IEEE/CVF Conference on Computer Vision and Pattern Recognition},
  year = {2021},
  pages = {10492--10502},
  url = {https://arxiv.org/abs/2103.04098}
}

@misc{insightfacemodelzoo,
  author = {{InsightFace}},
  title = {{InsightFace Model Zoo}},
  howpublished = {\url{https://github.com/deepinsight/insightface/tree/master/model_zoo}}
}

@inproceedings{cormack2009reciprocal,
author = {Cormack, Gordon and Clarke, Charles and Büttcher, Stefan},
year = {2009},
month = {07},
pages = {758-759},
title = {Reciprocal Rank Fusion outperforms Condorcet and Individual Rank Learning Methods},
journal = {Proc. 32nd Int. ACM SIGIR Conf. Res. Dev. Inf. Retrieval},
doi = {10.1145/1571941.1572114}
}

@article{fawcett2006introduction,
  author    = {Tom Fawcett},
  title     = {An introduction to {ROC} analysis},
  journal   = {Pattern Recognition Letters},
  volume    = {27},
  number    = {8},
  pages     = {861--874},
  year      = {2006}
}

@article{fisher1936use,
  author    = {Ronald A. Fisher},
  title     = {The use of multiple measurements in taxonomic problems},
  journal   = {Annals of Eugenics},
  volume    = {7},
  number    = {2},
  pages     = {179--188},
  year      = {1936}
}

@book{hosmer2013applied,
  author    = {David W. Hosmer and Stanley Lemeshow and Rodney X. Sturdivant},
  title     = {Applied Logistic Regression},
  edition   = {3rd},
  publisher = {John Wiley \& Sons},
  year      = {2013}
}
\bibliographystyle{iclr2027_conference}

\appendix

\setcounter{figure}{0}
\setcounter{table}{0}
\renewcommand{\thefigure}{A\arabic{figure}}
\renewcommand{\thetable}{A\arabic{table}}

\section{Additional Implementation Details}

\subsection{Recognition models and score processing}
\label{app:model_pools}
\label{app:score_normalization}
\label{app:composition}

\paragraph{Model configurations.}
The evaluated model combinations are specified in Tables~\ref{tab:ccvid_full}--\ref{tab:ltcc_full}. For the video benchmarks, AdaFace~\citep{adaface2022} uses ViT-Base~\citep{dosovitskiy2021vit} trained on WebFace4M~\citep{zhu2021webface260m}, and CAL~\citep{cal2022} uses a dataset-specific ResNet-50 checkpoint. We evaluate BigGait~\citep{biggait2024} on CCVID~\citep{ccvid2022}, AGRL~\citep{agrl} (ResNet-50 trained on MEVID~\citep{mevid2023}) on MEVID, and AIM~\citep{aim2023} on LTCC~\citep{ltcc2020}. On CCVID, InsightFace uses the released \texttt{buffalo\_l/w600k\_r50} ArcFace~\citep{arcface2019} model trained on WebFace600K~\citep{insightfacemodelzoo}. For CLIP-ReID~\citep{clipreid2023}, we use ViT-B/16~\citep{dosovitskiy2021vit} checkpoints trained on MSMT17~\citep{wei2018ptgan}, Market-1501~\citep{zheng2015market}, and DukeMTMC-reID~\citep{gou2017dukemtmc}, evaluated both with and without LoRA~\citep{lora2022} adaptation on the target benchmark's training split. Each contributes as a separate model in the combinations that include it.

\paragraph{Similarity scores of each model.}
Frame embeddings are mean-pooled within each tracklet; BigGait~\citep{biggait2024} and AGRL~\citep{agrl} provide sequence embeddings. Face embeddings use detected and aligned frames. On CCVID and MEVID~\citep{ccvid2022, mevid2023}, the embedding-based fusion experiments compute cosine similarity between unit-normalized embeddings for every model, including BigGait~\citep{biggait2024}. This applies to both query--gallery scoring and gallery--gallery profile estimation. On LTCC~\citep{ltcc2020}, query--gallery scores for AdaFace~\citep{adaface2022}, CAL~\citep{cal2022}, and AIM~\citep{aim2023} come from the released QME~\citep{qme2025} score matrices; CLIP-ReID~\citep{clipreid2023} scores and all gallery--gallery scores are computed from embeddings using cosine similarity. Larger scores indicate stronger matches.


\subsection{Additional Results}

\begin{figure}[t]
\centering
\includegraphics[width=.5\linewidth]{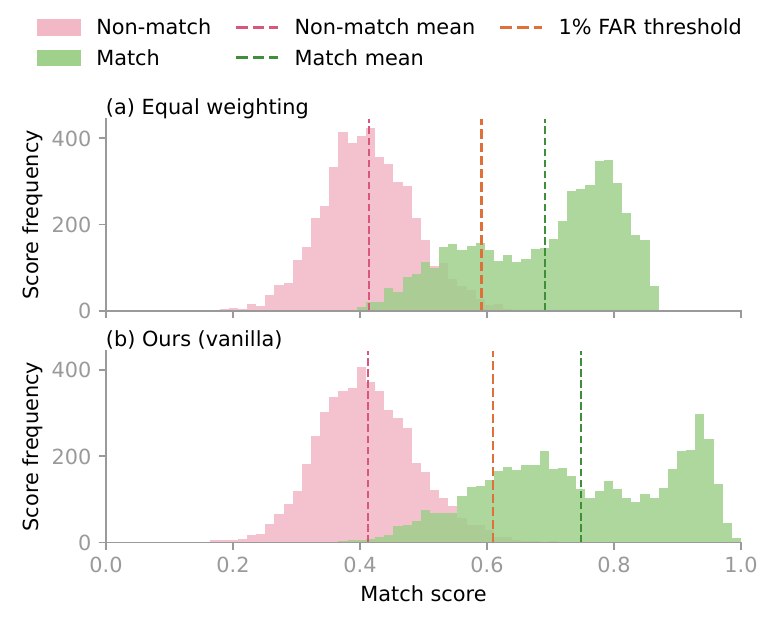}
\caption{\textbf{Separation of match and non-match score distributions.} Dataset: CCVID; model pool: AdaFace, BigGait, and CAL. Dashed lines indicate distribution means and the $1\%$ FAR threshold. Our identity-conditioned fusion markedly enlarges the margin between genuine matches and impostors by driving match scores rightward, effectively suppressing distribution overlap under strict verification thresholds.}
\label{fig:score_dist}
\end{figure}

\paragraph{Score distribution analysis.}
As demonstrated in \Figref{fig:score_dist}, our identity-conditioned fusion improves distributional separability over equal weighting on CCVID~\citep{ccvid2022}. By prioritizing each identity's most discriminative representations, our method elevates genuine match confidence. Because the two distributions overlap far less, fewer true matches are rejected when operating under a strict 1\% FAR threshold.

\paragraph{Weight combination strategies.}
\begin{table}[t]
\centering
\caption{\textbf{Ablation on weight combination rules for joint fusion.} Dataset: CCVID; Models pool: AdaFace, BigGait, CAL, and zero-shot CLIP-ReID. We compare different strategies for composing query-conditioned weights $\QueryFusionWeight{\query}{\modelIdx}$ and identity-specific weights $\FusionWeight{\idIdx}{\modelIdx}$, keeping each branch's weights fixed. $\lambda$ denotes the interpolation weight for the identity branch. \textbf{Bold} marks the best result among all combination strategies.}
\label{tab:joint_fusion}
\begingroup
\footnotesize
\setlength{\tabcolsep}{6pt}
\renewcommand{\arraystretch}{1.12}
\begin{tabular}{lcccc}
\toprule
Combination Rule & Rank-1 & mAP & TAR@1\%FAR & FNIR@1\%FPIR ($\downarrow$) \\
\midrule
Query only & 93.77 & 91.96 & 82.40 & 10.38 \\
Identity only & 93.53 & 93.10 & 85.36 & 11.23 \\
\midrule
Product (Default) & \textbf{94.60} & \textbf{93.99} & 85.60 & \textbf{9.96} \\
Weighted sum ($\lambda=0.25$) & 94.00 & 92.94 & 84.45 & 10.27 \\
Weighted sum ($\lambda=0.50$) & 93.89 & 93.38 & 85.94 & 10.51 \\
Weighted sum ($\lambda=0.75$) & 93.64 & 93.31 & 86.08 & 10.72 \\
Geometric mean & 93.64 & 93.40 & 85.50 & 10.57 \\
RRF ($k=60$) & 93.53 & 92.61 & \textbf{86.40} & 68.09 \\
\bottomrule
\end{tabular}
\endgroup
\end{table}
We ablate alternative rules for composing the query-conditioned weights $\QueryFusionWeight{\query}{\modelIdx}$ and identity-conditioned weights $\FusionWeight{\idIdx}{\modelIdx}$, keeping each branch's weights fixed. Specifically, the additive baselines linearly interpolate weights via $\lambda \FusionWeight{\idIdx}{\modelIdx} + (1-\lambda) \QueryFusionWeight{\query}{\modelIdx}$, geometric mean computes $(\FusionWeight{\idIdx}{\modelIdx} \QueryFusionWeight{\query}{\modelIdx})^{1/2}$, and RRF~\citep{cormack2009reciprocal} combines retrieval ranks instead of scores. Our product rule adopts element-wise multiplication $\FusionWeight{\idIdx}{\modelIdx} \cdot \QueryFusionWeight{\query}{\modelIdx}$ followed by normalization over available experts. As shown in \Tabref{tab:joint_fusion}, integrating both branches yields performance gains. Our multiplicative product outperforms additive interpolation across varying mixing weights $\lambda$, as well as geometric mean averaging, achieving the best Rank-1 ($94.60\%$), mAP ($93.99\%$), and open-set FNIR ($9.96\%$). Crucially, multiplicative gating acts as a parameter-free filter that naturally activates an expert only when query quality and identity reliability align. Although RRF achieves competitive TAR, its rank-transformed scores discard continuous similarity distributions, severely disrupting per-identity score calibration and causing open-set FNIR to degrade catastrophically to $68.09\%$. We therefore adopt the product rule as our default composition strategy in \Secref{sec:method:composition}.

\paragraph{Choice of per-identity separability metric.}
\begin{table}[t]
\centering
\caption{\textbf{Ablation on per-identity separability metrics.} Comparison of different gallery-derived separation scores and learned per-identity weighting baselines ($^\ddagger$), evaluated across four CCVID and three MEVID expert pools. FNIR and TAR are measured at $1\%$ FPIR and $1\%$ FAR, respectively. Bold indicates the best result per column.}
\label{tab:dprime_ablation}
\begingroup
\footnotesize
\setlength{\tabcolsep}{3pt}
\renewcommand{\arraystretch}{1.12}
\begin{tabular}{lcccccccc}
\toprule
 & \multicolumn{4}{c}{CCVID} & \multicolumn{4}{c}{MEVID} \\
\cmidrule(lr){2-5}\cmidrule(lr){6-9}
Approach & R-1 $\uparrow$ & mAP $\uparrow$ & TAR $\uparrow$ & FNIR $\downarrow$ & R-1 $\uparrow$ & mAP $\uparrow$ & TAR $\uparrow$ & FNIR $\downarrow$ \\
\midrule
Equal weights (no identity score) & $92.03$ & $90.62$ & $80.72$ & $15.96$ & $62.13$ & $36.62$ & $40.37$ & $58.92$ \\
\midrule
$d'$ (Default) & $\mathbf{94.21}$ & $\mathbf{92.62}$ & $\mathbf{84.36}$ & $10.60$ & $\mathbf{64.03}$ & $38.10$ & $41.75$ & $57.57$ \\
Mean genuine similarity & $93.65$ & $91.76$ & $83.09$ & $\mathbf{9.67}$ & $61.50$ & $36.10$ & $40.12$ & $59.97$ \\
Genuine $-$ impostor mean & $93.94$ & $91.95$ & $83.34$ & $9.74$ & $61.71$ & $36.44$ & $40.23$ & $59.82$ \\
AUC & $92.18$ & $90.76$ & $82.66$ & $16.46$ & $63.40$ & $37.89$ & $41.57$ & $57.84$ \\
$1-$EER & $92.24$ & $90.87$ & $83.07$ & $16.41$ & $63.08$ & $37.82$ & $41.44$ & $57.79$ \\
\midrule
LDA weights$^\ddagger$ & $89.24$ & $87.95$ & $81.99$ & $24.56$ & $62.34$ & $38.47$ & $\mathbf{42.74}$ & $\mathbf{57.52}$ \\
Logistic-regression weights$^\ddagger$ & $91.28$ & $89.56$ & $80.39$ & $20.09$ & $62.24$ & $\mathbf{38.60}$ & $42.01$ & $58.09$ \\
\bottomrule
\end{tabular}
\endgroup
\end{table}

We ablate the gallery-derived metric used to measure model distinctiveness per identity in \Tabref{tab:dprime_ablation}.
Starting from equal weights, we evaluate several training-free alternatives to $d'$.
Specifically, \emph{Mean genuine similarity} averages similarity scores within the same identity, while \emph{Genuine $-$ impostor mean} measures the gap between genuine and impostor averages.
We also evaluate threshold-independent separation using receiver operating characteristics, namely \emph{AUC} and \emph{$1-$EER}~\citep{fawcett2006introduction}.
Each score is normalized to match the global mean and variance of $d'$ before mapping to weights.
Finally, we test two learned per-identity baselines, linear discriminant analysis (LDA)~\citep{fisher1936use} and logistic regression~\citep{hosmer2013applied}, which fit model weights directly on gallery pairs with sample-size shrinkage toward the global solution.
Across both benchmarks, $d'$ yields the most reliable overall performance, demonstrating that accounting for score spread provides a stronger basis for weight estimation than other methods.

\paragraph{Alternative admission rule for gallery enrichment.}
\label{app:enriched}
\begin{figure}[t]
\centering
\includegraphics[width=\linewidth]{figs/enriched_legend}\\[2pt]
\begin{subfigure}[b]{0.333\linewidth}\centering\includegraphics[width=\linewidth]{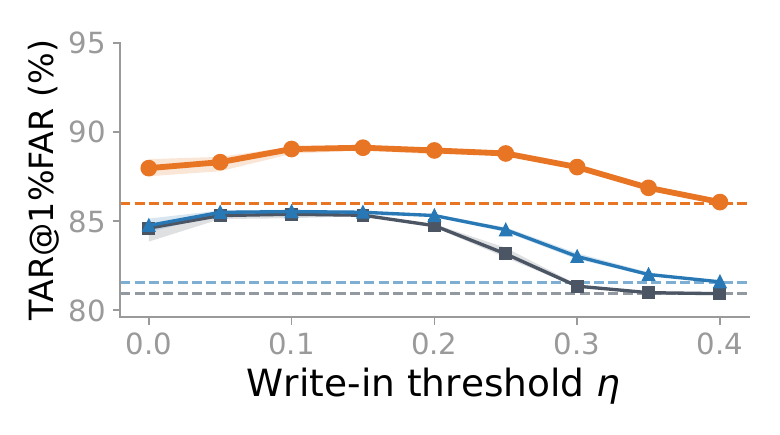}\caption{}\end{subfigure}\hfill
\begin{subfigure}[b]{0.333\linewidth}\centering\includegraphics[width=\linewidth]{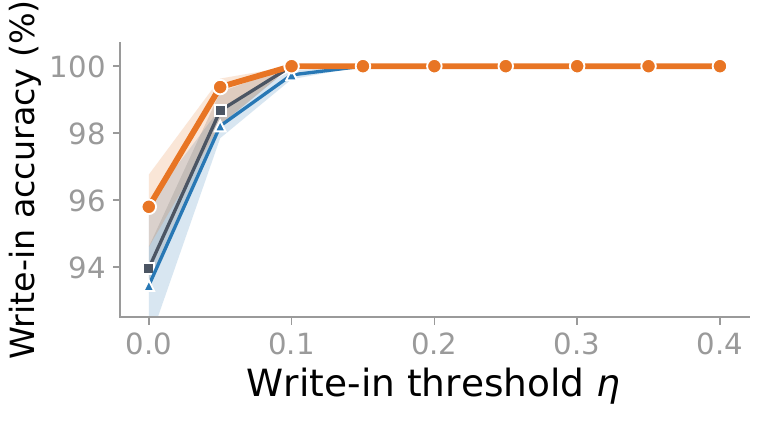}\caption{}\end{subfigure}\hfill
\begin{subfigure}[b]{0.333\linewidth}\centering\includegraphics[width=\linewidth]{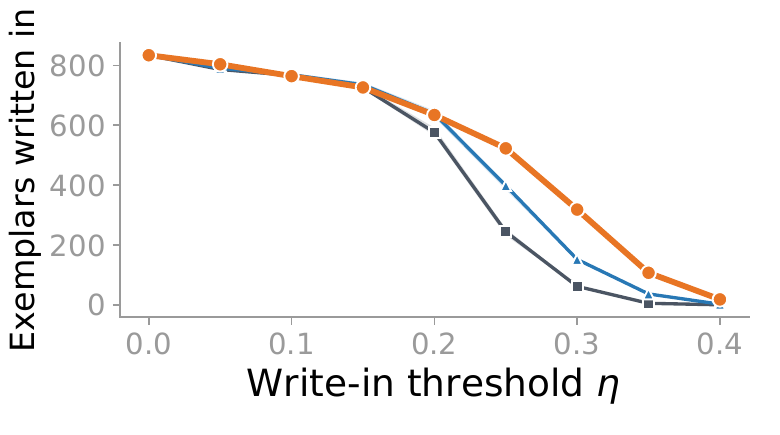}\caption{}\end{subfigure}
\caption{\textbf{Robustness to gallery enrichment with a margin-based admission rule.} Queries are enrolled when the gap between the top two candidate scores satisfies $\idScore{\query}{\matchedIdx} - \max_{\idIdx \neq \matchedIdx} \idScore{\query}{\idIdx} > \eta$. Across all subplots, curves and shaded areas indicate the mean and spread over 9 arrival orders, and dashed lines show static-gallery baselines: (a) TAR@$1\%$FAR, (b) pseudo-label precision of admitted queries, and (c) total admitted queries ($\eta = 0$ admits all queries). Ours (vanilla) consistently maintains its performance advantage over equal weighting and QME at every threshold.}
\label{fig:enriched_gap}
\end{figure}
While \Secref{sec:analysis:enriched} employs an absolute top-score threshold, we further verify whether our conclusions hold under an alternative, margin-based admission criterion. Under this rule, a query is admitted only when its top prediction leads the runner-up candidate by a confidence margin $\eta$, \ie $\idScore{\query}{\matchedIdx} - \max_{\idIdx \neq \matchedIdx} \idScore{\query}{\idIdx} > \eta$. Following the identical protocol on CCVID across nine arrival sequences, the resulting trends closely mirror those in the main text, as shown in \Figref{fig:enriched_gap}. 
All methods benefit from moderate enrichment before converging to their static-gallery baselines. Ours (vanilla) consistently outperforms equal weighting and QME across all thresholds; remarkably, its lowest mean TAR ($86.1\%$) still surpasses the peak performance of either baseline ($85.5\%$). The underlying enrollment dynamics in \Figref{fig:enriched_gap}(b) and (c) account for this advantage: even at $\eta = 0$ where all queries enter the gallery, Ours (vanilla) incurs lower label noise ($95.8\%$ accuracy vs.\ $\le 94.0\%$ for baselines). As $\eta$ increases, it consistently admits substantially more clean exemplars than the baselines under the same threshold. This confirms that identity-specific weighting sharpens margin separation between true matches and competing candidates, safely scaling up gallery enrichment without introducing detrimental noise.

\begin{table}[t]
\centering
\caption{\textbf{Sensitivity to hyperparameters.} Analysis of interpolation weight $\alpha$ and softmax temperature $\tau$ on CCVID using a model pool of AdaFace, BigGait, CAL, and InsightFace. FNIR and TAR are reported at $1\%$ FPIR and $1\%$ FAR, respectively. Bold indicates the best result per column.}
\label{tab:sweeps}
\begingroup
\footnotesize
\setlength{\tabcolsep}{3.5pt}
\renewcommand{\arraystretch}{1.12}
\begin{minipage}[t]{0.48\linewidth}
\centering
\textbf{(a) Interpolation weight $\alpha$ ($\tau=0.5$)}\\[3pt]
\begin{tabular}{lcccc}
\toprule
$\alpha$ & R-1 $\uparrow$ & mAP $\uparrow$ & TAR $\uparrow$ & FNIR $\downarrow$ \\
\midrule
$0$ (equal) & $93.53$ & $90.90$ & $80.92$ & $10.02$ \\
$0.25$ & $95.20$ & $92.08$ & $84.51$ & $9.29$ \\
$0.5$ (default) & $\mathbf{95.56}$ & $\mathbf{92.33}$ & $85.27$ & $\mathbf{8.97}$ \\
$0.75$ & $94.84$ & $91.21$ & $\mathbf{86.32}$ & 9.02 \\
$1$ & $93.05$ & $89.29$ & $85.33$ & $14.24$ \\
\bottomrule
\end{tabular}
\end{minipage}
\hfill
\begin{minipage}[t]{0.48\linewidth}
\centering
\textbf{(b) Softmax temperature $\tau$ ($\alpha=0.5$)}\\[3pt]
\begin{tabular}{lcccc}
\toprule
$\tau$ & R-1 $\uparrow$ & mAP $\uparrow$ & TAR $\uparrow$ & FNIR $\downarrow$ \\
\midrule
$0.1$ & $94.84$ & $91.54$ & $86.18$ & $9.85$ \\
$0.25$ & $95.32$ & $92.01$ & $\mathbf{86.34}$ & $9.45$ \\
$0.5$ (default) & $\mathbf{95.56}$ & $\mathbf{92.33}$ & $85.27$ & $\mathbf{8.97}$ \\
$1$ & $95.44$ & $92.23$ & $85.98$ & $9.20$ \\
$2$ & $95.44$ & $92.22$ & $84.73$ & $9.18$ \\
$4$ & $95.32$ & $91.98$ & $84.21$ & $9.35$ \\
\bottomrule
\end{tabular}
\end{minipage}
\endgroup
\end{table}
\paragraph{Sensitivity to hyperparameters.}
\Tabref{tab:sweeps} analyzes the effect of the interpolation weight $\alpha$ and softmax temperature $\tau$ using an expert pool comprising AdaFace, BigGait, CAL, and InsightFace on CCVID.
As shown in \Tabref{tab:sweeps}a, performance improves markedly from equal weighting ($\alpha=0$) across intermediate values of $\alpha$, peaking around $\alpha=0.5$.
Discarding the prior completely ($\alpha=1$) degrades verification, confirming that anchoring per-identity weights to a uniform baseline prevents overfitting to sparse gallery samples.
For the temperature parameter (\Tabref{tab:sweeps}b), recognition remains stable across a wide range of values ($0.25 \le \tau \le 2$), with $\tau=0.5$ providing a solid trade-off between closed-set retrieval and open-set verification without requiring aggressive tuning.

\paragraph{Extended evaluations across model pools.}
\label{app:extended_eval}
Tables~\ref{tab:ccvid_full}, \ref{tab:mevid_full}, and \ref{tab:ltcc_full} report full results across all model pools on CCVID, MEVID, and LTCC~\citep{ccvid2022, mevid2023, ltcc2020}. Across all benchmarks, \ourslab{vanilla} consistently outperforms single-model baselines and classical score fusion rules without requiring any parameter training. When integrated with query-adaptive methods (\ourslab{QME} and \ourslab{FusionAgent}), it achieves top-tier performance across diverse pool configurations. Notably, identity-specific weighting yields the most pronounced gains in open-set identification, substantially cutting FNIR@$1\%$FPIR over static fusion baselines and demonstrating strong robustness against severe appearance changes.

\begin{table}[!t]
\caption{\textbf{Extended multi-model performance on CCVID.} \emph{Combination} specifies the model pool shared by each block (\eFace~AdaFace, \eCAL~CAL, \eGait~BigGait,
\eIns~InsightFace/ArcFace~\citep{arcface2019}, and \eReID~zero-shot and \eReIDL~LoRA-adapted
CLIP-ReID \citep{clipreid2023,lora2022}). \best{Best} and \secondbest{second-best} results are marked within each model pool.}
\label{tab:ccvid_full}
\begin{center}
\small
\setlength{\tabcolsep}{3.5pt}
\tightrules
\renewcommand{\arraystretch}{1.05}
\setlength{\combrowheight}{10.15pt}
\resizebox*{!}{0.82\textheight}{%
\begin{tabular}{clcccc}
\toprule
\emph{Combination} & Method & Rank-1 & mAP & TAR@1\%FAR & FNIR@1\%FPIR ($\downarrow$) \\
\midrule
\comb{\eFace} & AdaFace \citep{adaface2022} & 94.0 & 87.9 & 75.7 & 13.0\pmm3.5 \\
\comb{\eCAL} & CAL \citep{cal2022} & 81.4 & 74.7 & 66.3 & 52.8\pmm13.3 \\
\comb{\eGait} & BigGait \citep{biggait2024} & 76.7 & 61.0 & 49.7 & 71.1\pmm6.1 \\
\comb{\eIns} & InsightFace/ArcFace \citep{arcface2019} & 94.1 & 88.2 & 78.9 & 18.1\pmm3.1 \\
\comb{\eReID} & CLIP-ReID \citep{clipreid2023} & 64.1 & 63.0 & 67.9 & 63.1\pmm2.5 \\
\comb{\eReIDL} & LoRA-adapted CLIP-ReID \citep{lora2022} & 85.4 & 84.3 & 80.3 & 42.8\pmm7.5 \\
\midrule
 & Min-Fusion \citep{jain2005score} & 87.1 & 79.2 & 62.4 & 48.5\pmm8.7 \\
 & Max-Fusion \citep{jain2005score} & 89.9 & 89.3 & 73.4 & 23.0\pmm10.1 \\
 & Z-score \citep{snelick2003multimodal} & 92.2 & 90.6 & 73.9 & 15.1\pmm1.5 \\
 & Min-max \citep{snelick2003multimodal} & 91.8 & 90.9 & 73.9 & 15.4\pmm2.0 \\
 & RHE \citep{he2010performance} & 91.7 & 90.2 & 73.1 & 16.6\pmm2.5 \\
 & Asym-AO \citep{herbadji2020combining} & 92.0 & 91.0 & 73.8 & 15.5\pmm1.8 \\
 & Farsight \citep{liu2025person} & 92.0 & 91.2 & 73.9 & 13.9\pmm1.1 \\
 & Weighted-sum \citep{park2021weighted} & 93.5 & 90.8 & 75.8 & 12.6\pmm1.1 \\
 & BSSF \citep{teng2022optimized} & 93.6 & 90.7 & 75.5 & 12.9\pmm1.6 \\
 & QME \citep{qme2025} & 94.1 & 90.8 & 76.2 & 12.3\pmm1.4 \\
 & FusionAgent \citep{fusionagent2026} & 93.4 & \secondbest{92.6} & \secondbest{85.9} & \secondbest{10.1\pmm1.5} \\
\ourrow & \ourslab{vanilla} & \best{95.3} & 92.1 & 81.0 & 11.0\pmm1.6 \\
\ourrow & \ourslab{QME} & \secondbest{95.2} & 91.9 & 79.3 & 10.3\pmm1.5 \\
\ourrow \combblock{14}{\eFace\eCAL\eGait} & \ourslab{FusionAgent} & 94.6 & \best{93.7} & \best{87.0} & \best{9.4\pmm1.2} \\
\midrule
 & Min-Fusion \citep{jain2005score} & 94.4 & 88.6 & 79.6 & 15.1\pmm2.6 \\
 & Max-Fusion \citep{jain2005score} & 84.2 & 69.2 & 57.5 & 51.2\pmm8.3 \\
 & Z-score \citep{snelick2003multimodal} & 93.5 & 90.8 & 80.9 & 13.2\pmm2.5 \\
 & Min-max \citep{snelick2003multimodal} & 93.4 & 90.9 & 78.8 & 13.5\pmm2.5 \\
 & RHE \citep{he2010performance} & 93.4 & 90.9 & 79.0 & 13.5\pmm2.3 \\
 & Asym-AO \citep{herbadji2020combining} & 93.3 & 90.7 & 78.3 & 13.5\pmm2.5 \\
 & Farsight \citep{liu2025person} & 93.6 & 90.7 & 79.2 & 13.3\pmm2.0 \\
 & Weighted-sum \citep{park2021weighted} & 94.2 & 89.8 & 81.5 & 13.5\pmm2.1 \\
 & BSSF \citep{teng2022optimized} & 93.5 & 90.6 & 82.8 & 15.0\pmm3.9 \\
 & QME \citep{qme2025} & 93.0 & 91.5 & 78.9 & 11.2\pmm1.2 \\
\ourrow & \ourslab{vanilla} & \best{95.4} & \secondbest{92.2} & \best{86.0} & \best{9.2\pmm1.4} \\
\ourrow \combblock{12}{\eFace\eCAL\eGait\eIns} & \ourslab{QME} & \secondbest{94.7} & \best{92.7} & \secondbest{84.2} & \secondbest{9.6\pmm1.3} \\
\midrule
 & Min-Fusion \citep{jain2005score} & \secondbest{94.1} & 87.9 & 74.4 & 15.4\pmm1.1 \\
 & Max-Fusion \citep{jain2005score} & 79.9 & 73.5 & 75.3 & 55.7\pmm4.9 \\
 & Z-score \citep{snelick2003multimodal} & 89.6 & 89.3 & 82.7 & 35.8\pmm7.0 \\
 & Min-max \citep{snelick2003multimodal} & 88.6 & 88.2 & 82.5 & 37.8\pmm6.0 \\
 & RHE \citep{he2010performance} & 88.7 & 88.2 & 82.5 & 38.9\pmm6.6 \\
 & Asym-AO \citep{herbadji2020combining} & 90.9 & 90.8 & 81.3 & 31.6\pmm6.2 \\
 & Farsight \citep{liu2025person} & 91.7 & 91.6 & 80.9 & 15.3\pmm3.7 \\
 & Weighted-sum \citep{park2021weighted} & 94.0 & 92.4 & 84.4 & 14.0\pmm2.9 \\
 & BSSF \citep{teng2022optimized} & 93.9 & 92.3 & \secondbest{85.1} & 17.0\pmm4.8 \\
 & QME \citep{qme2025} & 92.3 & 92.3 & 81.6 & 11.2\pmm1.5 \\
\ourrow & \ourslab{vanilla} & \best{94.6} & \best{94.0} & \best{85.6} & \best{10.0\pmm1.4} \\
\ourrow \combblock{12}{\eFace\eCAL\eGait\eReID} & \ourslab{QME} & 93.2 & \secondbest{93.0} & 83.3 & \secondbest{11.0\pmm1.4} \\
\midrule
 & Min-Fusion \citep{jain2005score} & 91.6 & 87.6 & 75.7 & 20.6\pmm3.9 \\
 & Max-Fusion \citep{jain2005score} & 84.3 & 74.4 & 65.5 & 49.2\pmm7.1 \\
 & Z-score \citep{snelick2003multimodal} & 91.8 & 92.0 & 83.6 & 18.3\pmm4.7 \\
 & Min-max \citep{snelick2003multimodal} & 91.8 & 91.9 & 83.0 & 18.8\pmm4.6 \\
 & RHE \citep{he2010performance} & 91.8 & 91.9 & 83.0 & 18.9\pmm4.5 \\
 & Asym-AO \citep{herbadji2020combining} & 91.8 & 92.4 & 82.4 & 16.5\pmm4.6 \\
 & Farsight \citep{liu2025person} & 91.8 & 91.8 & 83.2 & 19.5\pmm4.2 \\
 & Weighted-sum \citep{park2021weighted} & \best{93.2} & \secondbest{92.7} & \best{85.8} & 12.1\pmm1.4 \\
 & BSSF \citep{teng2022optimized} & \secondbest{92.8} & 92.5 & \best{85.8} & 14.5\pmm4.1 \\
 & QME \citep{qme2025} & 91.7 & 91.7 & 82.9 & 14.7\pmm2.7 \\
\ourrow & \ourslab{vanilla} & 92.6 & \best{93.0} & \secondbest{85.1} & \best{10.9\pmm1.2} \\
\ourrow \combblock{12}{\eFace\eCAL\eGait\eReIDL} & \ourslab{QME} & 91.7 & 92.3 & 83.2 & \secondbest{11.0\pmm1.2} \\
\bottomrule
\end{tabular}%
}
\end{center}
\end{table}
\begin{table}[!t]
\caption{\textbf{Extended multi-model evaluation on MEVID.} \emph{Combination} specifies the model pool shared by each block (\eFace~AdaFace, \eCAL~CAL, \eGait~AGRL, and \eReID~zero-shot and \eReIDL~LoRA-adapted CLIP-ReID \citep{clipreid2023,lora2022}). \best{Best} and \secondbest{second-best} results are marked within each block.}
\label{tab:mevid_full}
\begin{center}
\small
\setlength{\tabcolsep}{3.5pt}
\tightrules
\begin{tabular}{clcccc}
\toprule
\emph{Combination} & Method & Rank-1 & mAP & TAR@1\%FAR & FNIR@1\%FPIR ($\downarrow$) \\
\midrule
\comb{\eFace} & AdaFace \citep{adaface2022} & 25.0 & 8.1 & 5.4 & 98.8\pmm1.2 \\
\comb{\eCAL} & CAL \citep{cal2022} & 52.5 & 27.1 & 34.7 & 67.8\pmm7.3 \\
\comb{\eGait} & AGRL \citep{agrl} & 51.9 & 25.5 & 30.7 & 69.4\pmm8.9 \\
\comb{\eReID} & CLIP-ReID \citep{clipreid2023} & 63.0 & 33.0 & 33.6 & 68.3\pmm3.3 \\
\comb{\eReIDL} & LoRA-adapted CLIP-ReID \citep{lora2022} & 69.9 & 45.1 & 47.7 & 49.7\pmm4.3 \\
\midrule
 & Min-Fusion \citep{jain2005score} & 46.8 & 21.2 & 28.0 & 70.4\pmm8.0 \\
 & Max-Fusion \citep{jain2005score} & 33.2 & 14.9 & 8.3 & 97.4\pmm1.6 \\
 & Z-score \citep{snelick2003multimodal} & 54.1 & 27.4 & 30.7 & 66.5\pmm7.0 \\
 & Min-max \citep{snelick2003multimodal} & 52.8 & 24.7 & 25.0 & 71.3\pmm6.1 \\
 & RHE \citep{he2010performance} & 52.8 & 24.8 & 25.3 & 71.2\pmm6.2 \\
 & Asym-AO \citep{herbadji2020combining} & 52.5 & 22.9 & 23.6 & 71.7\pmm6.4 \\
 & Farsight \citep{liu2025person} & 53.8 & 25.4 & 26.6 & 69.8\pmm7.1 \\
 & Weighted-sum \citep{park2021weighted} & 53.5 & 29.0 & 34.8 & 64.8\pmm8.9 \\
 & BSSF \citep{teng2022optimized} & 54.4 & 28.7 & 33.4 & 63.8\pmm9.4 \\
 & QME \citep{qme2025} & \secondbest{55.7} & 28.2 & 32.9 & 64.6\pmm8.2 \\
 & FusionAgent \citep{fusionagent2026} & 54.7 & 28.7 & 34.9 & \secondbest{58.6\pmm7.4} \\
\ourrow & \ourslab{vanilla} & 55.4 & \secondbest{31.0} & \secondbest{36.2} & 66.4\pmm13.6 \\
\ourrow & \ourslab{QME} & 54.7 & \secondbest{31.0} & 35.5 & 59.0\pmm12.6 \\
\ourrow \combblock{14}{\eFace\eCAL\eGait} & \ourslab{FusionAgent} & \best{57.0} & \best{31.2} & \best{37.3} & \best{52.0\pmm15.5} \\
\midrule
 & Min-Fusion \citep{jain2005score} & 47.8 & 21.7 & 28.0 & 70.4\pmm8.0 \\
 & Max-Fusion \citep{jain2005score} & 62.0 & 35.1 & 35.9 & 63.7\pmm3.2 \\
 & Z-score \citep{snelick2003multimodal} & 62.7 & 34.5 & 37.9 & 56.4\pmm8.8 \\
 & Min-max \citep{snelick2003multimodal} & 62.0 & 33.2 & 34.9 & 59.8\pmm8.6 \\
 & RHE \citep{he2010performance} & 61.1 & 32.6 & 34.2 & 61.3\pmm8.5 \\
 & Asym-AO \citep{herbadji2020combining} & 58.2 & 28.8 & 28.6 & 62.0\pmm7.7 \\
 & Farsight \citep{liu2025person} & 56.0 & 29.3 & 30.7 & 65.6\pmm6.8 \\
 & Weighted-sum \citep{park2021weighted} & 61.1 & 34.0 & 39.4 & 58.4\pmm6.7 \\
 & BSSF \citep{teng2022optimized} & 62.0 & 36.1 & 39.6 & 57.5\pmm6.2 \\
 & QME \citep{qme2025} & 63.6 & 36.0 & 40.1 & \secondbest{55.6\pmm8.4} \\
\ourrow & \ourslab{vanilla} & \best{65.8} & \best{37.7} & \best{40.9} & 57.3\pmm16.8 \\
\ourrow \combblock{12}{\eFace\eCAL\eGait\eReID} & \ourslab{QME} & \secondbest{64.2} & \secondbest{37.0} & \secondbest{40.7} & \best{50.1\pmm16.7} \\
\midrule
 & Min-Fusion \citep{jain2005score} & 47.8 & 21.7 & 28.0 & 70.4\pmm8.0 \\
 & Max-Fusion \citep{jain2005score} & \secondbest{68.4} & 43.9 & 45.9 & 55.8\pmm8.1 \\
 & Z-score \citep{snelick2003multimodal} & 66.8 & 41.1 & 44.9 & 52.1\pmm7.2 \\
 & Min-max \citep{snelick2003multimodal} & 66.1 & 40.3 & 42.6 & 54.8\pmm8.1 \\
 & RHE \citep{he2010performance} & 65.2 & 39.0 & 41.2 & 57.0\pmm8.7 \\
 & Asym-AO \citep{herbadji2020combining} & 62.3 & 35.0 & 31.3 & 57.8\pmm8.0 \\
 & Farsight \citep{liu2025person} & 61.7 & 36.6 & 38.7 & 58.9\pmm9.1 \\
 & Weighted-sum \citep{park2021weighted} & 65.5 & 40.4 & 45.2 & 52.6\pmm5.6 \\
 & QME \citep{qme2025} & 65.5 & 42.1 & 46.3 & 51.2\pmm6.1 \\
\ourrow & \ourslab{vanilla} & \best{70.9} & \best{45.6} & \best{48.2} & \secondbest{49.0\pmm18.5} \\
\ourrow \combblock{11}{\eFace\eCAL\eGait\eReIDL} & \ourslab{QME} & 67.7 & \secondbest{44.2} & \secondbest{47.6} & \best{48.2\pmm16.8} \\
\bottomrule
\end{tabular}
\end{center}
\end{table}
\begin{table}[!t]
\caption{\textbf{Extended multi-model performance on LTCC.} \emph{Combination} specifies the model
pool shared by each block (\eFace~AdaFace, \eCAL~CAL, \eAIM~AIM, and \eReID~zero-shot
and \eReIDL~LoRA-adapted CLIP-ReID \citep{clipreid2023,lora2022}). \best{Best} and \secondbest{second-best} results are marked within each model pool.}
\label{tab:ltcc_full}
\vspace{-2pt}
\begin{center}
\small
\setlength{\tabcolsep}{3.5pt}
\tightrules
\begin{tabular}{clcccc}
\toprule
\emph{Combination} & Method & Rank-1 & mAP & TAR@1\%FAR & FNIR@1\%FPIR ($\downarrow$) \\
\midrule
\comb{\eFace} & AdaFace \citep{adaface2022} & 18.5 & 5.9 & 2.4 & 99.8\pmm0.2 \\
\comb{\eCAL} & CAL \citep{cal2022} & 74.4 & 40.6 & 36.7 & 59.7\pmm7.3 \\
\comb{\eAIM} & AIM \citep{aim2023} & 74.8 & 40.9 & 37.0 & 66.2\pmm7.5 \\
\comb{\eReID} & CLIP-ReID \citep{clipreid2023} & 70.7 & 33.7 & 26.5 & 47.8\pmm13.0 \\
\comb{\eReIDL} & LoRA-adapted CLIP-ReID \citep{lora2022} & 76.7 & 40.7 & 31.9 & 42.0\pmm10.3 \\
\midrule
 & Min-Fusion \citep{jain2005score} & 37.1 & 12.9 & 12.4 & 81.9\pmm6.0 \\
 & Max-Fusion \citep{jain2005score} & 62.5 & 33.3 & 16.8 & 94.8\pmm4.7 \\
 & Z-score \citep{snelick2003multimodal} & 73.0 & 37.5 & 30.4 & 68.7\pmm9.2 \\
 & Min-max \citep{snelick2003multimodal} & 73.2 & 38.1 & 31.9 & 75.1\pmm9.2 \\
 & RHE \citep{he2010performance} & 70.4 & 34.2 & 21.5 & 78.0\pmm10.0 \\
 & Asym-AO \citep{herbadji2020combining} & 71.2 & 32.9 & 19.1 & 76.3\pmm8.9 \\
 & Farsight \citep{liu2025person} & 73.2 & 37.8 & 31.3 & 72.4\pmm8.6 \\
 & Weighted-sum \citep{park2021weighted} & 75.1 & \best{42.5} & \best{38.1} & 59.5\pmm9.4 \\
 & BSSF \citep{teng2022optimized} & 75.1 & \best{42.5} & \secondbest{38.0} & 60.0\pmm9.6 \\
 & QME \citep{qme2025} & 73.8 & 39.6 & 35.0 & 64.3\pmm8.0 \\
 & FusionAgent \citep{fusionagent2026} & \secondbest{75.5} & 41.0 & 37.0 & 50.0\pmm8.5 \\
\ourrow & \ourslab{vanilla} & 70.8 & 40.3 & 35.1 & 52.0\pmm12.7 \\
\ourrow & \ourslab{QME} & \best{75.7} & \secondbest{42.4} & 37.9 & \best{26.4\pmm10.2} \\
\ourrow \combblock{14}{\eFace\eCAL\eAIM} & \ourslab{FusionAgent} & \secondbest{75.5} & 41.0 & 35.2 & \secondbest{33.6\pmm14.0} \\
\midrule
 & Min-Fusion \citep{jain2005score} & 37.1 & 12.9 & 12.4 & 81.9\pmm6.0 \\
 & Max-Fusion \citep{jain2005score} & 72.4 & 35.6 & 26.5 & 49.4\pmm21.2 \\
 & Z-score \citep{snelick2003multimodal} & 76.7 & 42.8 & 36.3 & 44.4\pmm9.2 \\
 & Min-max \citep{snelick2003multimodal} & 77.5 & 43.2 & 36.5 & 51.0\pmm9.9 \\
 & RHE \citep{he2010performance} & 75.9 & 41.0 & 33.8 & 64.5\pmm9.4 \\
 & Asym-AO \citep{herbadji2020combining} & 74.4 & 31.3 & 27.6 & 60.6\pmm7.2 \\
 & Farsight \citep{liu2025person} & 76.3 & 42.1 & 35.9 & 64.2\pmm9.6 \\
 & Weighted-sum \citep{park2021weighted} & \secondbest{77.7} & \best{44.9} & \best{39.8} & 54.5\pmm11.6 \\
 & BSSF \citep{teng2022optimized} & 77.5 & \secondbest{44.5} & \secondbest{39.4} & 56.8\pmm11.1 \\
 & QME \citep{qme2025} & \best{77.9} & 43.6 & 37.6 & 59.8\pmm7.0 \\
\ourrow & \ourslab{vanilla} & 76.9 & \best{44.9} & 38.3 & \secondbest{35.6\pmm16.6} \\
\ourrow \combblock{12}{\eFace\eCAL\eAIM\eReID} & \ourslab{QME} & 77.5 & 43.9 & 37.7 & \best{28.5\pmm20.0} \\
\midrule
 & Min-Fusion \citep{jain2005score} & 42.0 & 15.0 & 13.1 & 81.9\pmm6.0 \\
 & Max-Fusion \citep{jain2005score} & 76.1 & 41.6 & 32.1 & 32.6\pmm11.0 \\
 & Z-score \citep{snelick2003multimodal} & 79.1 & 47.8 & 40.3 & 48.3\pmm13.3 \\
 & Min-max \citep{snelick2003multimodal} & 79.9 & 48.7 & 40.6 & 53.6\pmm15.3 \\
 & RHE \citep{he2010performance} & 78.3 & 46.3 & 38.4 & 64.9\pmm14.7 \\
 & Asym-AO \citep{herbadji2020combining} & 76.3 & 38.8 & 32.2 & 63.0\pmm9.8 \\
 & Farsight \citep{liu2025person} & 80.3 & 48.6 & 40.4 & 49.9\pmm14.6 \\
 & Weighted-sum \citep{park2021weighted} & 79.3 & 48.4 & \best{41.9} & 51.8\pmm9.1 \\
 & BSSF \citep{teng2022optimized} & 79.9 & \secondbest{49.3} & \best{41.9} & 50.9\pmm10.0 \\
 & QME \citep{qme2025} & \best{80.7} & 48.5 & 41.3 & 60.7\pmm8.5 \\
\ourrow & \ourslab{vanilla} & \secondbest{80.5} & \best{49.6} & \secondbest{41.4} & \secondbest{31.1\pmm19.0} \\
\ourrow \combblock{12}{\eFace\eCAL\eAIM\eReIDL} & \ourslab{QME} & \secondbest{80.5} & 48.7 & 41.3 & \best{29.1\pmm21.2} \\
\bottomrule
\end{tabular}
\end{center}
\end{table}

\end{document}